\documentclass{article}

\PassOptionsToPackage{numbers, compress}{natbib}
\usepackage[preprint]{conference}

\usepackage[utf8]{inputenc} %
\usepackage[T1]{fontenc}    %
\usepackage{hyperref}       %
\usepackage{url}            %
\usepackage{booktabs}       %
\usepackage{amsfonts}       %
\usepackage{nicefrac}       %
\usepackage{microtype}      %
\usepackage{xcolor}         %
\usepackage{kotex}

\usepackage{amsmath}
\usepackage{bbm}
\usepackage{multirow}
\usepackage{graphicx}
\usepackage{tabularx} 
\usepackage{algorithm}
\usepackage{algpseudocode}
\usepackage{comment}
\usepackage{wrapfig}
\usepackage{afterpage}

\usepackage{tcolorbox}
\tcbuselibrary{skins}

\newcommand{\dataname}{\texttt{V-RAGBench}}
\newcommand{\algname}{\texttt{CARVE}}

\DeclareMathOperator*{\argmax}{arg\,max}

\definecolor{absgray}{RGB}{242,243,245}
\definecolor{metablue}{RGB}{0,102,204}

\hypersetup{colorlinks,linkcolor={blue},citecolor={blue},urlcolor={metablue}}

\title{Rethinking RAG in Long Videos: \\What to Retrieve and How to Use It?}

\begin{document}

\newcommand{\customabstractpage}{
\begin{tcolorbox}[
    enhanced,
    colback=absgray,
    colframe=absgray,
    boxrule=0pt,
    arc=8pt,
    left=3mm,
    right=3mm,
    top=3mm,
    bottom=3mm
]

{\Large\bfseries
Rethinking RAG in Long Videos: \\What to Retrieve and How to Use It?
\par}

\vspace{3mm}

Yuho Lee$^{1}$, Jisu Shin$^{1}$, Nicole Hee-Yeon Kim$^{1}$, Jihwan Bang$^{2}$, Juntae Lee$^{2}$, Kyuwoong Hwang$^{2}$, Fatih Porikli$^{2}$ Hwanjun Song$^{1}$\par

\vspace{1mm}

$^{1}$KAIST, $^{2}$Qualcomm AI Research$^{\dagger}$, Qualcomm Korea YH, Seoul, Republic of Korea \par

\vspace{1mm}

\vspace{4mm}

\noindent
Retrieval-augmented generation is extending beyond text to long videos, where query-relevant chunks can be represented across multiple modalities and temporal granularities. Progress in this setting, VideoRAG, is limited by two gaps: existing benchmarks allow queries to be answered without the video, obscuring retrieval errors, and prior methods apply a single modality-granularity configuration per query, ignoring chunk-level variability. We address both by introducing \dataname{}, a benchmark of $\langle$query, evidence chunk, answer$\rangle$ triplets over hour-scale videos, in which each answer depends on a unique evidence chunk, enabling decoupled evaluation of retrieval and generation, and \algname{}, a training-free method that runs parallel retrievers across configurations and uses chunk-adaptive reranking to select a winning configuration for each chunk, which is then carried into generation. On \dataname{}, \algname{} outperforms {eight} recent VideoRAG baselines on both stages, with the evidence it supplies to the generator interleaving multiple configurations rather than sharing one. Its gains hold across egocentric and third-person long videos and extend to an expanded configuration space.

\vspace{4mm}

\noindent
\begin{minipage}[t]{0.65\textwidth}
{\small
\textbf{Date:} June 11, 2026\par
\textbf{Correspondence:} Hwanjun Song
({\color{metablue}\href{mailto:songhwanjun@kaist.ac.kr}{songhwanjun@kaist.ac.kr}})\par
\textbf{First Authors (equal contribution):}\par
Yuho Lee ({\color{metablue}\href{mailto:yuholee@kaist.ac.kr}{yuholee@kaist.ac.kr}})\par
Jisu Shin ({\color{metablue}\href{mailto:jisu3389@kaist.ac.kr}{jisu3389@kaist.ac.kr}})\par
Nicole Hee-Yeon Kim ({\color{metablue}\href{mailto:nicole2kim@gmail.com}{nicole2kim@gmail.com}})\par
}
\end{minipage}
\hfill
\begin{minipage}[t]{0.23\textwidth}
\vspace*{-0.1cm}
\raggedleft
\includegraphics[width=1.0\linewidth]{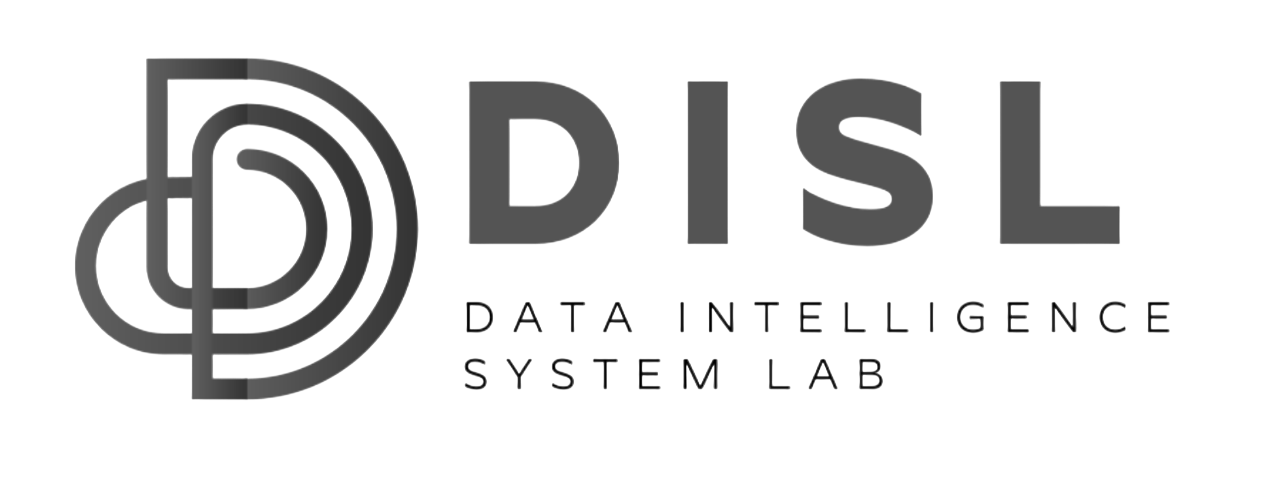}
\vspace{2mm}
\includegraphics[width=1.0\linewidth]{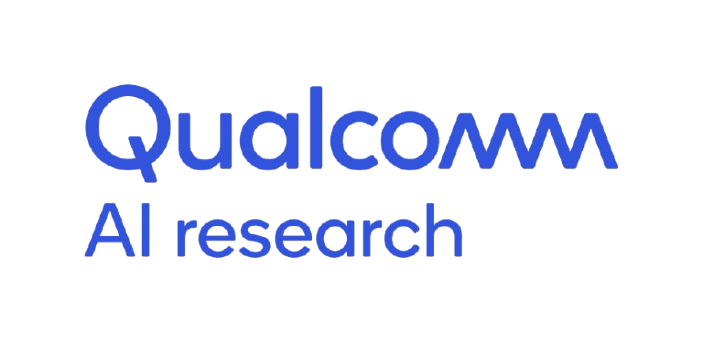}
\end{minipage}

{\fontsize{7}{9}\selectfont\noindent $^{\dagger}$Qualcomm AI Research is an initiative of Qualcomm Technologies, Inc.\par}

\end{tcolorbox}
}

\thispagestyle{empty}

\vspace*{-1.3cm}

\customabstractpage

\vspace*{-0.3cm}
\section{Introduction}
\label{sec:intro}
\vspace*{-0.2cm}

Retrieval-augmented generation (RAG) grounds large language models (LLMs) in external knowledge, improving factual reliability \citep{choi2025word2passage, singh2025agentic, zhao2026retrieval}. Its frontier is now extending from text corpora to \emph{long-form video} \citep{jeong2025videorag, zhang2025deep, xue2025adavideorag, yeo2025worldmm}, such as first-person recordings from wearable devices \citep{bock2024wear, yang2025egolife} and video histories that agentic systems operate over \citep{chen2025lvagent, rege2026agentic}. We refer to RAG over long videos as \texttt{VideoRAG}. Unlike a text passage, a video chunk has no single canonical form: it can be matched through different \emph{modalities} \citep{zhang2024simple, maaz2024videochatgpt, kahatapitiya2025language}, from raw visual features to textual abstractions, and at different \emph{temporal granularities} \citep{lin2024multi, hu2025mllm}, from individual frames to whole clips. The problem thus moves beyond what to retrieve to "which modality and granularity to retrieve each chunk from, and how to use them across retrieval and generation?"

Despite this growing interest \citep{lin2024mm, ma2025drvideo, sun2026egograph}, current benchmark datasets have not kept pace. While hour-scale source videos are available through Ego4D \citep{grauman2022ego4d} and EgoLife \citep{yang2025egolife}, the queries paired with them are scarce and were not designed for \texttt{VideoRAG}. In particular, recent work \citep{lim2026video} shows that over half of widely used video QA samples can be answered without the video at all, relying on linguistic priors, world knowledge, or static cues. Such queries undermine \texttt{VideoRAG} evaluation, since a system can reach correct answers even when retrieval surfaces irrelevant chunks, so \emph{high generation accuracy does not imply successful retrieval}, and final accuracy cannot serve as a proxy for the retrieval stage. Consequently, how modality and temporal granularity should be used across retrieval and generation cannot be identified, as each design choice is invisible behind the final accuracy.

\begin{figure}[t!]
    \centering
    \includegraphics[width=1\textwidth]{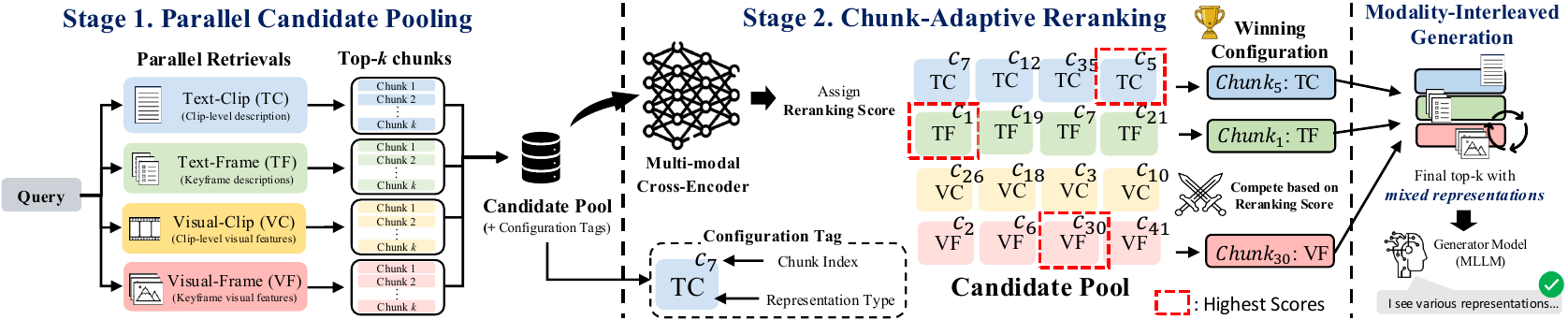}
    \vspace*{-0.5cm}
    \caption{The overview of \algname{}: Stage 1 builds a candidate pool via chunk-wise parallel retrieval, while Stage 2 performs chunk-adaptive reranking. The final top-$k$ evidence with their winning configuration is passed to the generator in a modality-interleaved form.}
    \label{fig:method}
    \vspace*{-0.45cm}
\end{figure}

The same gap appears on the methodological side. Recent work proposes techniques for \texttt{VideoRAG}, such as guided query refinement \citep{wang2025active, xue2025adavideorag}, modality fusion \citep{luo2025video, yang2026graph}, and query decomposition \citep{xu2025vrag}, while others place retrieval inside iterative agentic loops \citep{ma2025drvideo, yeo2025worldmm}. Although retrieval is the component on which all these frameworks rest, its validation is reduced to downstream QA accuracy of the full pipeline. This contrasts with text-centric RAG, where retrieval is studied as a first-class object and audited through stage-wise metrics on query-evidence pairs \citep{formal2021splade, choi2025word2passage}.

We address this gap by introducing \dataname{}, a benchmark designed to make retrieval and generation in \texttt{VideoRAG} measurable both in isolation and in their interaction. \dataname{} consists of $2{,}100$ $\langle${query}, {evidence chunk}, {answer}$\rangle$ triplets over $216$ hour-scale videos ($1$--$9$ hours) drawn from Ego4D \citep{grauman2022ego4d} and EgoLife \citep{yang2025egolife}. It jointly enforces three properties that no existing dataset satisfies together: each query is (i) \emph{tied to non-recurring evidence}, so that the underlying event does not repeat elsewhere in the video; (ii) \emph{visually-grounded}, so that the answer is not implied by the query itself or recoverable from parametric knowledge alone; and (iii) \emph{evidence-localized}, so that the answer cannot be reconstructed from cues that recur across non-evidence chunks. These properties, verified by exhaustive human audits, make generation causally dependent on retrieval, enabling a stage-wise view of \texttt{VideoRAG}.

With \dataname{}, we shift the question of \texttt{VideoRAG} from improving QA accuracy to which representation of a chunk best supports each stage, retrieval and generation. Each chunk admits four configurations\footnote{Other retrieval signals, such as video knowledge graphs \citep{shen2025vgent, dixit2026seg} or metadata (\emph{e.g.}, explicit temporal clue) \citep{ren2024timechat, guo2025vtg}, can aid retrieval but are not consistently available. We focus on content-based representations, which are universally accessible.} from two axes: modality (visual vs. textual-summary embeddings) and temporal granularity (frame vs. clip). Existing strategies \citep{hsu2025dat, jeong2025videorag, han2026cova, ren2026videorag} apply one configuration, or a fixed fusion, uniformly to every chunk of a query---a \emph{query-level} decision. However, this is suboptimal as the best configuration depends on the chunk's content, not only on the query; a visually salient moment is best surfaced by visual embeddings, a semantically rich one by its textual summary, and likewise for granularity. The question is thus at which level this decision should be made. We answer it at the chunk level with \algname{} (\underline{C}hunk-\underline{A}ware \underline{R}eranking for \underline{V}ideo \underline{E}vidence; Figure \ref{fig:method}), which re-purposes a multi-modal cross-encoder, a standard reranking tool in text-RAG \citep{li2024llatrieval, abdallah2025rerankarena, wang2025infogain}, to assign each chunk its winning configuration and feed it to the generator under that configuration. Being training-free and adding only parallel retrieval and reranking to the pipeline, \algname{} lets us isolate the effect of the decision level and serves as a drop-in retrieval module for existing pipelines, including agentic ones, as shown in Appendix \ref{app:add_agentic}.

Concretely, \algname{} operates in two stages. Parallel retrieval runs one retriever per configuration, each contributing its top-$k$ chunks to a shared candidate pool. Chunk-adaptive reranking then re-scores each candidate with a multi-modal cross-encoder under its retrieving configuration (\emph{e.g.}, query--text relevance for textual retrievals, query--visual for visual ones), yielding a final ranking that interleaves chunks from different configurations, each tagged with its winning one. These top-$k$ chunks serve both stages. Retrieval outputs them with their tags, and generation receives each chunk under its winning configuration, so the representation fed to the generator varies per chunk. Importantly, our experiments reveal that this chunk-level selection improves both retrieval and generation, surpassing existing VideoRAG methods and even trained query-level routers.

Our experiments on \dataname{} reveal four key findings: (1) \algname{} significantly outperforms {eight} baseline methods on both retrieval and generation, a comparison enabled by \dataname{}'s stage-wise measurability; (2) a modality--granularity ablation reveals large performance gaps across configurations at both stages, with no single configuration uniformly optimal, motivating chunk-level rather than query-level decisions; (3) \algname{}'s chunk-level decisions distribute fairly evenly across the four configurations, showing that per-chunk diversity is genuine rather than collapsing onto a single dominant choice; and (4) at generation, \algname{} even surpasses trained query-level routers that learns to pick one configuration per query, without any additional training.

\vspace*{-0.1cm}
\section{Related Work}
\vspace*{-0.1cm}

\paragraph{Retrieval-Augmented Generation in Text.}
RAG retrieves query-relevant evidence from an external text corpus and conditions the generator on it \citep{lewis2020retrieval, jiang2023active, arslan2024survey, choi2025word2passage}. Since retrieval quality largely dictates answer quality, progress has centered on three fronts. Retrievers have advanced from sparse and dense models \citep{karpukhin2020dense, formal2021splade, yan2024corrective} to hybrid pipelines that combine both \citep{mandikal2024sparse, sawarkar2024blended, hsu2025dat}. Queries are reformulated through rewriting \citep{ma2023query, wang2025maferw, zhao2026rewritegen} or expansion \citep{azad2019query, zhang2024exploring, choi2025word2passage} to make retrieval intent explicit. Reranking re-scores first-stage candidates with a cross-encoder to improve precision \citep{li2024llatrieval, yu2024rankrag, abdallah2025rerankarena, wang2025infogain}. These techniques are developed for text passages, which have a single canonical form, and do not address which representation of a video chunk to retrieve from.

\textbf{Retrieval-Augmented Generation in Long Video.} Long-form video understanding has traditionally relied on query-independent representations of the whole video, such as sparse frame sampling \citep{lin2023videollava,chasmai2025moment,  yao2025generative}, visual-token compression \citep{zhang2023videollama, cheng2024videollama2, huang2025prunevid, wang2025dynamic}, or full-video textualization \citep{liu2021video, zhang2024simple, lee2025video, li2025videochat}, which can dilute query-relevant evidence as length grows. \texttt{VideoRAG} instead retrieves query-relevant chunks before generation. A video chunk, however, is inherently multimodal \citep{ maaz2024videochatgpt, zhang2024simple, kahatapitiya2025language}, expressible through visual frames or derived textual abstractions, and readable at temporal granularities from individual frames to segments \citep{lin2024multi, hu2025mllm}. 

Existing work explores these axes separately. On modality, \citet{jeong2025videorag} retrieves chunks with multimodal embeddings and \citet{luo2025video} augments retrieval with visually-aligned auxiliary text such as OCR and ASR. On granularity, hierarchical and tree-structured representations index the video at multiple temporal scales \citep{wang2024videotree, ren2026videorag}. A further line replaces one-shot retrieval with iterative agentic probing that repeatedly fetches frames or segments \citep{wang2024videoagent, chen2025lvagent, xue2025adavideorag, yeo2025worldmm}. Across these methods, the modality--granularity choice, or a fixed fusion of the options, is made once per query, and evaluation rests mainly on downstream QA accuracy. Our work treats the choice as a chunk-level decision and introduces a benchmark that measures retrieval and generation separately.

\textbf{Video Benchmark Datasets for QA.} Video QA benchmarks have progressed from short third-person clips \citep{xu2017videoqa, lei2018tvqa, yu2019activitynetqa} toward longer and increasingly egocentric videos \citep{grauman2022ego4d, barmann2022qaego4d, mangalam2023egoschema, patel2025advancing, yang2025egolife}, a natural target of \texttt{VideoRAG} as wearable devices and personal agents accumulate hour-scale first-person video \citep{zhou2025x, rege2026agentic}. Yet existing benchmarks remain limited in video length, \emph{e.g.}, EgoSchema (3-minute clips) \citep{mangalam2023egoschema}, QaEgo4D (short Ego4D intervals) \citep{barmann2022qaego4d}, Video-MME ($\le$60 min) \citep{fu2024videomme}, and LongVideoBench ($\le$1 hour) \citep{wu2024longvideobench}. Ego4D \citep{grauman2022ego4d} and EgoLife \citep{yang2025egolife} provide hour-scale source video, but their queries are scarce and serve other goals. Ego4D's Episodic Memory benchmark evaluates retrieval alone, and EgoLifeQA targets memory aggregation across days. Neither guarantees the three properties that \texttt{VideoRAG} evaluation requires, namely (i) non-recurring evidence, (ii) visual grounding, and (iii) evidence localization, which tie correct answers to the labeled evidence rather than to spurious cues. CG-Bench \citep{chen2024cgbench} and MM-Lifelong \citep{chen2026mmlifelong} check shortcut answerability and clue grounding, addressing (ii) but not (i) and (iii). \dataname{} jointly enforces all three on hour-scale video.

\vspace*{-0.1cm}
\section{Formulating VideoRAG along Two Design Axes}
\label{sec:preliminaries}
\vspace*{-0.1cm}

We formalize \texttt{VideoRAG} as a retrieve-then-generate process over long videos. Prior studies \citep{ren2026videorag, yeo2025worldmm} establish this pipeline but leave two design choices underspecified. First, a chunk admits multiple representations across modality (visual vs. textual) and granularity (frame vs. clip), yet these are combined at fixed scales rather than treated as decision variables. Second, the representation used for retrieval and the one used for generation are tied to a single choice by default, and their relationship has received little attention.

Following standard text RAG practice \citep{arslan2024survey, singh2025agentic}, a \texttt{VideoRAG} system partitions a long video $V$ into non-overlapping, fixed-interval temporal chunks $V=\{v_1, \dots, v_{|V|}\}$ that serve as retrieval units. We default to $2$-minute chunks, matching the empirical $2$--$3$ minute span of bounded semantic content in long-form video \citep{islam2024videorecap}. Unlike a text passage, each chunk $v$ admits multiple representations $\phi_{m,g}(v)$ along two axes, a modality $m\in\{\mathrm{vis}, \mathrm{text}\}$ (visual features vs.\ textual abstractions) and a temporal granularity $g\in\{\mathrm{frame}, \mathrm{clip}\}$ (frame-level details vs.\ clip-level summaries).\footnote{We instantiate $\mathcal{M}$ and $\mathcal{G}$ with two values each, giving four $(m, g)$ configurations for ease of exposition. The formulation and \algname{} accommodate additional modalities and granularities without any change, and we extend the space to six configurations with ASR and OCR in Appendix \ref{app:add_config_scaling}.} Given a query $q$, a retriever $\mathcal{R}$ matches $q$ against chunk representations to select $k$ candidate chunks $V_q\subseteq V$, and an MLLM-based generator $\mathcal{G}$ produces an answer from those chunks under a chosen representation:

\begin{equation}
\widehat{a} = \mathcal{G}\!\left(q,\; \{\phi_{m_g, g_g}(v)\}_{v\in V_q}\right)~~{\rm where}~~V_q = \mathcal{R}\!\left(q,\; \{\phi_{m_r, g_r}(v)\}_{v\in V}\right),
\qquad
\label{eq:rag}
\end{equation}
$(m_r, g_r)$ and $(m_g, g_g)$ are the configurations chosen for retrieval and generation; and $\phi_{m,g}(v)$ is the representation of chunk $v$ under modality $m$ and granularity $g$. The goal is twofold: $V_q$ should contain the chunk that supports the answer, and $\widehat{a}$ should match the reference answer $a$. Notably, whereas a text chunk has a single canonical form, a video chunk forces \texttt{VideoRAG} to decide \emph{which $(m, g)$ to use for retrieval and which for generation}, and these decisions need not coincide.

We now describe how the two design axes are instantiated in this work. Full implementation details (\emph{e.g.}, decoding hyperparameters, temperature) are deferred to Appendix \ref{app:implimentation}, and the prompt templates used are provided in Appendix \ref{app:prompts}.

\textbf{Chunk Modality.~} The modality axis $m\in\{\mathrm{vis}, \mathrm{text}\}$ exposes a chunk $v$ through two complementary channels. The \emph{visual} modality encodes $v$ as raw visual features from an MLLM-based encoder (Qwen3-VL-Embedding-2B \citep{qwen3vlembedding} by default), preserving fine-grained appearance and motion cues that resist verbalization. In contrast, the \emph{textual} modality prompts an MLLM (Qwen3-VL-8B) to summarize $v$ in natural language and embeds the summary with the same encoder, surfacing semantic structure (\emph{e.g.}, what is happening, where, and with whom) at the cost of pixel-level detail. Neither is universally preferable, as which is more useful depends on the query and chunk content.

\textbf{Chunk Granularity.~} Even within a fixed modality, the granularity axis $g\in\{\mathrm{frame}, \mathrm{clip}\}$ exposes a chunk $v$ at two temporal scales. The \emph{frame}-level retains per-frame detail through a small set of keyframes, obtained by clustering all frame embeddings in the chunk with $k$-Means++ \citep{arthur2007k} ($n=5$ by default) and taking the centroids, which cover distinct visual states rather than uniform time samples. The \emph{clip}-level treats $v$ as a single unit. Each is realized in a modality-specific way. Under the visual modality, the frame-level mean-pools the keyframe embeddings, while the clip-level encodes the entire chunk into one embedding. Under the textual modality, the same MLLM produces a summary capped at $512$ tokens, applied per keyframe (concatenated before embedding) for the frame-level, while over the entire chunk for the clip-level. The cap accommodates the detailed descriptions used in prior multimodal summarization work \citep{chen2024sharegpt4video, islam2024videorecap} while keeping verbose outputs from dominating the context. Neither granularity is universally preferable, as frame-level forms capture transient details but lose the surrounding flow, while clip-level forms retain temporal context but blur such details.

\textbf{Scope and Objective.} With $(m, g)$ now made explicit, two questions follow: \emph{how} it should be decided and \emph{at what level}. In other words, for each chunk $v$, which representation $\phi_{m,g}(v)$ should be passed to the retriever and the generator? The common choice fixes one $(m, g)$ per query and applies it uniformly across chunks \citep{jeong2025videorag, ren2026videorag}, and agentic variants \citep{yeo2025worldmm} issue multiple retrievals per query but commit each call to a single $(m, g)$. The most useful $(m, g)$, however, may differ across chunks even within one query. For instance, identifying an object visible in only one keyframe is best handled by a frame-level visual embedding, while describing how an activity unfolds over the chunk is better expressed by a clip-level textual summary. Given a query $q$, we therefore decide $(m, g)$ \emph{per chunk} and carry the choice consistently into both retrieval and generation, keeping the standard retrieve-then-generate pipeline intact.

\vspace*{-0.25cm}
\section{Construction of V-RAGBench}
\label{sec:benchmark}
\vspace*{-0.2cm}

Diagnosing how the two design axes interact at retrieval and generation requires a benchmark in which the two stages are measurable both in isolation and their interaction, which in turn requires every query to be tied to a uniquely sufficient evidence chunk. We thus construct \dataname{}, a benchmark of $\langle$query, evidence chunk, answer$\rangle$ triplets built from hour-scale videos via four stages (see Figure \ref{fig:vragbench-pipeline}): source selection, event extraction and deduplication, query generation, and filtering.

\textbf{Source Video Selection.~} We draw source videos from egocentric corpora. Unlike third-person long videos (\emph{e.g.}, movies, lectures) organized around a single narrative arc, first-person streams capture continuous, weakly structured daily activity in which the same actor revisits similar scenes and routines, so retrieval must separate the queried moment from many near-duplicate distractors, which makes it a demanding testbed for \texttt{VideoRAG}.\footnote{The benchmark is egocentric by construction, but the method is not. We also evaluate \algname{} on third-person long videos (LongVideoBench) and on lecture videos with ASR/OCR (Video-MMMU) in Appendices  \ref{app:add_longvideobench}--\ref{app:add_config_scaling}}. We use Ego4D \citep{grauman2022ego4d} and EgoLife \citep{yang2025egolife}, retaining only videos \emph{longer than one hour} and applying stratified sampling over length bins so that the corpus spans a wide range of durations. This yields $216$ videos in total, comprising $42$ EgoLife recordings (avg.\ $379.1$ min., $177.8$--$552.7$ min.) and $174$ Ego4D videos (avg.\ $86.0$ min., $60.0$--$423.6$ min.); see Table \ref{tab:statistics}. Retrieval is performed within each source video, so every query is grounded in a single long video. Domain and length statistics are in Appendix \ref{app:statistics}.

\textbf{Event Extraction and Deduplication.~} This stage converts each long video into a small set of distinct event chunks that form a non-redundant evidence pool for QA generation. For each video, we encode every sampled frame with EVA02-E-14-plus \citep{sun2023eva}, a CLIP-based vision encoder, and apply kernel temporal segmentation \citep{potapov2014category} on the frame embeddings to cut the video into semantically coherent, time-contiguous segments. The number of segments is capped at $150$ for the shorter Ego4D videos and $900$ for the day-long EgoLife recordings, with each segment constrained to $5$--$300$s. We then re-embed each segment with Qwen3-VL-Embedding-8B \citep{qwen3vlembedding} in video mode, cluster the embeddings with k-Means++ ($k\!=\!50$ for Ego4D, $k\!=\!300$ for EgoLife), and retain only the segment closest to each centroid. This suppresses recurring routines (\emph{e.g.}, repeated meal preparation) and leaves chunks that correspond to \emph{distinct events}, which is a prerequisite for QA generation, since queries grounded in recurring events would admit multiple sufficient evidence chunks. Such queries are natural in the real world but are excluded here for rigorous \texttt{VideoRAG} evaluation. Details and statistics of this stage are in Appendix \ref{app:event_stats}.

\textbf{Query Generation.} For each event chunk, we prompt Gemini-3-flash-preview to generate up to four candidate queries, each assigned to one of three predefined categories (Appendix \ref{app:catagories}); queries are generated only for categories the chunk visually supports. The prompt enforces \emph{contextual localization}, requiring every query to embed anchor information from its source event so that the query alone describes the situation enough to be matched against the right chunk. This lets \dataname{} dispense with direct temporal cues such as timestamps or when-phrases, so retrieval must rely purely on content (prompt and examples in Appendix \ref{app:prompts}). Each generation yields a triple $\langle q,\, [t_s, t_e],\, a \rangle$ of query, evidence time span, and an \emph{open-ended} answer rather than a multiple-choice option, so that downstream evaluation reflects the generator's ability rather than random guessing over a fixed choice set. This produces $67{,}370$ candidate $\langle\text{query},\, \text{evidence chunk},\, \text{answer}\rangle$ triples for filtering.

\textbf{Post-hoc Filtering.~} Event deduplication enforces uniqueness at the event level, but the generated queries must still be verified at the query level for \emph{visual grounding}, so that the answer is not implied by the query or by priors, and \emph{evidence localization}, so that the answer is not recoverable from non-evidence chunks.
We thus introduce five filters, ordered from cheapest to most expensive (see Table \ref{tab:filter_detail}), using models other than the query generator to limit shared biases except where video input length forces the same family. (1) \textit{Semantic Similarity} removes queries too close in Qwen3-VL-embedding-8B, with per-video caps of $50$ (Ego4D) and $300$ (EgoLife) so that no single video dominates; (2) \textit{Shortcut Bias} discards queries that GPT-5.2-chat answers correctly without visual context; (3) \textit{Answerability} removes queries that GPT-5.2-chat judges unanswerable from the source clip; (4) \textit{Empirical Answerability} requires Gemini-2.5-flash to answer correctly from the source clip, judged by an LLM-as-a-judge; and (5) \textit{Evidence Uniqueness} retrieves the top-$10$ chunks per query, \emph{excluding} the true evidence, and discards the query if Gemini-2.5-flash still answers correctly.

The surviving queries are thus (1) distinct, (2) free of parametric and commonsense shortcuts, (3, 4) grounded in sufficient evidence, and (5) not recoverable from any other chunk of the same video. Lastly, since the three query categories are unevenly populated, we subsample each to the smallest one ($n\!=\!700$), yielding $2{,}100$ queries ($1{,}800$ train, $300$ test).

\smallskip
\textbf{Human Verification.~} 
To validate the output of the filtering pipeline, Table \ref{tab:human_audit} reports human verification of answerability, evidence uniqueness, and query non-redundancy, where each item is labeled independently by two annotators.
For answer accuracy to reflect retrieval, the labeled chunk must suffice to answer the query and no other chunk may, so that a correct answer implies successful retrieval. Annotators confirm both. On $577$ QA pairs sampled to represent the full benchmark at a $95$\% confidence level, $96.9$\% are judged answerable from the labeled chunk (IAA $0.63$). Over every two-minute window of test videos, judged without any retriever, $97.1$\% of windows outside the labeled evidence are unanswerable ($95$\% CI $96.8$--$97.5$, IAA $0.96$). Annotators also find no duplicate among the $2{,}648$ same-video test query pairs (IAA $1.00$). See Appendix \ref{app:human_audit} for details.

\begin{table}[t!]
\caption{Human verification of \dataname{}, each item labeled independently by two annotators.}
\label{tab:human_audit}
\centering
\scriptsize
\setlength{\tabcolsep}{3pt}
\begin{tabular}{@{}cccc@{}}
\toprule
{Requirement} & {Property Verified} & {Items (Judgments)} & {Result (Inter-annotator agreement)} \\
\midrule
Answerability & Answerable from labeled evidence & 577 sampled QA pairs (1,154) & 96.9\% answerable ($\kappa$ 0.63) \\
Evidence Uniqueness & Non-recurring, localized evidence & 10,206 test window--query pairs (20,412) & 97.1\% unanswerable (AC1 0.96) \\
Query Redundancy & Query-level non-redundancy & 2,648 same-video test pairs (5,296) & 0 duplicates (AC1 1.00) \\
\bottomrule
\end{tabular}
\vspace*{-0.35cm}
\end{table}

\vspace*{-0.2cm}
\section{CARVE: Chunk-Aware Reranking for Video Evidence}
\label{sec:method}
\vspace*{-0.1cm}

\algname{} is a training-free instantiation of chunk-level configuration decision. It adds only parallel retrieval and a reranking step to the standard retrieve-then-generate pipeline, so its gain over query-level baselines is attributable to deciding $(m, g)$ per chunk and carrying that decision into generation. It operates in two stages, parallel candidate pooling and chunk-adaptive reranking, as in Figure \ref{fig:method}.

\smallskip
\textbf{Stage 1: Parallel Candidate Pooling.} Given a query $q$, this stage builds a \emph{candidate pool} by performing four parallel retrievals, one per configuration $(m, g)$ with $m \in \{\mathrm{vis}, \mathrm{text}\}$ and $g \in \{\mathrm{frame}, \mathrm{clip}\}$. The four embeddings $\phi_{m,g}(v)$ of each chunk $v$ in Eq.~(\ref{eq:rag}) are precomputed offline, so each retrieval only encodes the query $q$ and performs a nearest-neighbor lookup, incurring no chunk-encoding cost at inference regardless of how many configurations are queried. Adding configurations grows the candidate pool sublinearly, as analyzed in Appendix \ref{app:add_config_scaling}.

For each configuration $(m, g)$, we compute the dot-product similarity $\langle \cdot,\, \cdot \rangle$ between the query embedding $q$ and each chunk embedding $\phi_{m,g}(v)$, and keep the $k$ chunks with the highest scores as the top-$k$ retrieval list $C_{(m,g)}(q)$ for that configuration:
\begin{equation}
C_{(m,g)}(q) \;=\; \text{top-}k \text{ chunks } v \text{ ranked by } \langle q,\, \phi_{m,g}(v) \rangle.
\label{eq:per_config_retrieval}
\end{equation}
The four lists are then merged into a configuration-tagged candidate pool ${P}(q)$, where each retrieved chunk is paired with the configuration $(m, g)$ that retrieved the chunk $v$:
\begin{equation}
{P}(q) \;=\; \bigcup_{(m, g)} \Big\{\, \big(\,\underbrace{v}_{\text{chunk}},\; \underbrace{(m, g)}_{\text{tag}}\,\big) \;:\; v \in C_{(m,g)}(q) \,\Big\}.
\label{eq:pool}
\end{equation}
The same chunk $v$ may appear under multiple tags if it was retrieved by more than one configuration.
These tags are kept, since they are the per-configuration signal from which Stage 2 chooses a winning $(m^*, g^*)$ for each chunk, unlike late fusion or reciprocal-rank fusion, which merge them into one score and lose which configuration surfaced the chunk.

\textbf{Stage 2: Chunk-Adaptive Reranking.} Given the candidate pool $\mathcal{P}(q)$ from Stage 1, this stage selects the final top-$k$ chunks, while assigning each surviving chunk a single \emph{winning} configuration $(m^*, g^*)$. Since the chunk embeddings in $\mathcal{P}(q)$ live in different modalities depending on the tag---textual when $m = \mathrm{text}$, visual when $m = \mathrm{vis}$---we use a multi-modal cross-encoder $\mathrm{CE}(\cdot,\,\cdot)$\footnote{Temporal granularity $g$ does not change the modality of $\phi_{m,g}(v)$ since frame- and clip-level forms under the same modality share the same encoder branch and embedding space, differing only in temporal scope (Section~\ref{sec:preliminaries}).} that jointly attends to a textual query and either textual or visual chunk modality. We use Qwen3-VL-Reranker-2B as the default reranker, though performance remains consistent across rerankers, as in Appendix \ref{app:add_reranker_backbone}.

Importantly, each chunk $v$ in $\mathcal{P}(q)$ is re-scored \emph{only} under the tag $(m, g)$ that retrieved it, not under all four configurations. The rerank score of a tagged candidate $(v, (m, g)) \in \mathcal{P}(q)$ is computed: 
\begin{equation}
\tilde{s}\big(q;\, (v, (m, g))\big) \;=\; \mathrm{CE}\big(q,\, \phi_{m,g}(v)\big).
\label{eq:rerank_score}
\end{equation}
For each chunk $v$, we keep only its highest-scoring tag as its winning configuration:
\begin{equation}
(m^*_v,\, g^*_v) \;=\; \argmax_{(m, g) \,:\, (v, (m, g)) \in \mathcal{P}(q)} \tilde{s}\big(q;\, (v, (m, g))\big).
\label{eq:winner}
\end{equation}
$\mathrm{CE}$ scores are comparable across modalities, so chunks can be ranked together by their winning scores, yielding the final top-$k$ list ($V^*_q$) with each chunk tagged by its winning configuration:
\begin{equation}
V^*_q \;=\; \Big\{\, \big(\,v,\, (m^*_v, g^*_v)\,\big) \;:\; v \in \text{top-}k \text{ chunks ranked by } \tilde{s}\big(q;\, (v, (m^*_v, g^*_v))\big) \,\Big\}.
\label{eq:final_retrieval}
\end{equation}
This chunk-adaptive reranking improves retrieval by judging each chunk under its most discriminative configuration and letting chunks from different configurations compete on a comparable scale, so the top-$k$ naturally interleaves the strongest evidence across modalities and granularities.

\textbf{Modality-Interleaved Generation.} $V^*_q$ benefits not only retrieval but also generation. Unlike prior work that fixes a single $(m_g, g_g)$ for the whole query \citep{jeong2025videorag, ren2026videorag} or fuses several representations uniformly, we expose each chunk to the generator $\mathcal{G}$ \emph{only} through its winning representation $\phi_{m^*_v, g^*_v}(v)$:
\begin{equation}
\widehat{a} \;=\; \mathcal{G}\!\left(q,\; \big\{\phi_{m^*_v,\, g^*_v}(v) \;:\; (v, (m^*_v, g^*_v)) \in V^*_q\big\}\right).
\label{eq:gen}
\end{equation}
The generator context thus mixes representations chunk by chunk, with each chunk rendered under its winning configuration. Empirically, this propagation from retrieval to generation---not the retrieval gain alone---is where \algname{}'s strongest improvements appear. 

The simplicity of \algname{} is deliberate. Being training-free with a standard retrieval interface, it replaces the retrieval module of existing \texttt{VideoRAG} pipelines without modification, including agentic frameworks, and improves both stages there, as shown in Appendix \ref{app:add_agentic}.

\vspace*{-0.1cm}
\section{Evaluation on V-RAGBench}\label{sec:evaluation}
\vspace*{-0.1cm}

Our evaluation of \algname{} on the test set of \dataname{} consists of three parts. First, we compare \algname{} against recent \texttt{VideoRAG} methods on both retrieval and generation to demonstrate its superiority (Table \ref{tab:baseline_comparison}). Second, we study how modality and temporal granularity should be selected under our method, showing that \algname{}'s chunk-level choices outperform query-level alternatives (Table \ref{tab:hybrid_comparison}). Third, we further show the efficiency and effectiveness of chunk-level decisions via latency, selection distribution, and comparison against two trained query-level routers (Tables \ref{tab:rank_memory}--\ref{tab:routing_comparison}).

\textbf{Compared Methods.} We compare \algname{} against eight RAG methods applicable to \texttt{VideoRAG}, grouped into two categories. (i) \textit{System-level} methods apply a fixed policy regardless of the query, including VideoRAG-A \citep{jeong2025videorag}, a late fusion method that linearly interpolates text and visual scores; GQR \citep{uzan2026guided}, which refines query embeddings at test time via complementary retriever signals; and the multimodal encoders FreeRet \citep{zhu2026freeret} and GME \citep{zhang2025gme}, which retrieve in a single embedding space. (ii) \textit{Query-level} methods adapt the policy per query but apply the same decision to all chunks, including RRF \citep{rackauckas2024rag}, which expands a query into variants and merges their ranks by reciprocal rank fusion; DAT \citep{hsu2025dat}, a hybrid fusion method that weights dense and sparse retrievers via LLM-assessed reference scores; and VideoRAG-B \citep{luo2025video} and UniversalRAG(-LoRA) \citep{yeo2026universalrag}, which respectively decompose or route a query across auxiliary databases of different modalities and granularities, where (-LoRA) is fine-tuned on the \dataname{} training set. {These baselines comprehensively cover late fusion of per-modality scores (VideoRAG-A), reciprocal-rank fusion of ranked lists (RRF), and hybrid fusion with adaptive retriever weights (DAT).} For fairness, all baselines share the same chunk embeddings as \algname{} and differ only in how representations are selected or combined. See Appendix \ref{app:baseline_implementation} for implementation details.

\smallskip
\textbf{Evaluation Metrics.} We evaluate retrieval and generation separately. For retrieval, we report two metrics: \emph{nDCG@$k$}, which rewards methods that rank the annotated evidence higher; and \emph{Recall@$k$}, which measures whether the annotated evidence appears among the top-$k$ retrieved chunks. Since retrieval relies on deterministic embedding similarity, these metrics carry no run-to-run variance.

For generation, we report the aggregate \emph{pass rate}, defined as the fraction of queries whose generated answer is judged correct against the reference by an LLM-as-a-judge (Qwen3.6-35B-A3B), which is necessary since answers are in {open-ended} form rather than multiple choice (see the prompt in Figure \ref{fig:llmjudge}; {$99.0$\% of its decisions are upheld by human annotators in Appendix~\ref{app:judge})}. To assess generalization across generators, we evaluate each retrieval method with three backbones: \{Qwen3-VL-8B, Qwen3-VL-32B, and Gemma-4-26B\}. We fix $k=5$ throughout experiments, except in Appendix~\ref{app:add_topk} where we vary $k$ to show that \algname{} consistently outperforms baselines. Detailed definitions of all metrics are provided in Appendix \ref{app:metrics}.

\vspace*{-0.15cm}
\subsection{Main Result: Performance on Retrieval and Generation}\label{sec:result_1_main}
\vspace*{-0.1cm}

\begin{table}[t!]
\caption{Retrieval and generation performance on \dataname{}. Generation is reported as pass rate under three generator backbones. Retrieval is deterministic, so significance is tested on pass rate only, averaged over three runs (refer to Appendix \ref{app:add_stat} for details). $^{*}$: $p<0.05$ for \algname{} against the second-best method (Welch's $t$-test). The best values are marked in bold.}
\label{tab:baseline_comparison}
\centering
\scriptsize
\setlength{\tabcolsep}{3pt}
\begin{tabular}{@{}llccccc@{}}
\toprule
& & \multicolumn{2}{c}{{Retrieval Performance}} & \multicolumn{3}{c}{{Generation Performance (w. Three Backbones)}} \\
\cmidrule(lr){3-4} \cmidrule(l){5-7}
{Category} & {Method} & ~~~{Recall@5}~~~ & ~~~{nDCG@5}~~~ & {Qwen3-VL-8B} & {Qwen3-VL-32B} & {Gemma-4-26B} \\
\midrule
System-Level & VideoRAG-A \citep{jeong2025videorag} & 0.510 & 0.332 & 0.238 & 0.312 & 0.305 \\
& GQR~\citep{uzan2026guided} & 0.503 & 0.340 & 0.167 & 0.110 & 0.153 \\
& FreeRet~\citep{zhu2026freeret} & 0.263 & 0.187 & 0.227 & 0.200 & 0.160 \\
& GME ~\citep{zhang2025gme} & 0.413 & 0.253 & 0.253 & 0.282 & 0.212\\ \midrule
Query-Level & RRF \citep{rackauckas2024rag} & 0.463 & 0.298 & 0.180 & 0.130 & 0.187 \\
& DAT~\citep{hsu2025dat} & 0.460 & 0.312 & 0.223 & 0.147 & 0.200 \\
& VideoRAG-B \citep{luo2025video} & 0.487 & 0.325 & 0.312 & 0.303 & 0.236 \\
& UniversalRAG \citep{yeo2026universalrag} & 0.447 & 0.298 & 0.259 & 0.253 & 0.228 \\
& UniversalRAG-LoRA \citep{yeo2026universalrag} & 0.470 & 0.311 & 0.237 & 0.270 & 0.220 \\ \midrule
\textbf{Chunk-Level} & \textbf{\algname{} (Ours)} & \textbf{0.603} & \textbf{0.433} & ~~\textbf{0.352}$^{*}$ & ~~\textbf{0.368}$^{*}$ & ~~\textbf{0.318}$^{*}$ \\ \bottomrule
\end{tabular}
\vspace*{-0.35cm}
\end{table}

Table \ref{tab:baseline_comparison} compares \algname{} with four system-level and four query-level baselines on \dataname{}. On retrieval, \algname{} achieves the best Recall@5 ($0.603$) and nDCG@5 ($0.433$), exceeding the strongest baselines, VideoRAG-A at $0.510$ Recall@5 and GQR at $0.340$ nDCG@5. This retrieval gain translates into stronger generation on all three backbones. \algname{} reaches a pass rate of $0.318$--$0.368$, a $4.3$--$17.9$\% relative improvement over the second-best method with every margin statistically significant.
The chunk-level decision strengthens both stages: chunk-adaptive reranking surfaces the evidence chunk under its winning configuration, which is then carried forward as interleaved evidence for generation, improving answers on every backbone. The same trend holds beyond egocentric video. On the third-person LongVideoBench and on Video-MMMU lecture videos with ASR and OCR added, \algname{} again ranks first on all three backbones, as reported in Appendices \ref{app:add_longvideobench}--\ref{app:add_config_scaling}. It also stays first on an extended $500$-query set in Appendix \ref{app:add_stat}. Remaining failures are mostly cases where retrieval succeeds but the generator misorders the actions in time, which we detail in Appendix \ref{app:add_error_analysis}.

\begin{wraptable}{r}{0.5\textwidth}
\vspace*{-0.65cm}
\caption{Ablation over modality--granularity configurations in \algname{}.
}
\vspace*{0.1cm}
\label{tab:hybrid_comparison}
\scriptsize
\setlength{\tabcolsep}{2pt}
\renewcommand{\arraystretch}{1}
\begin{tabular}{@{}lcccc@{}}
\toprule
{Configuration} 
& \shortstack[c]{R@5}
& \shortstack[c]{nDCG@5}
& \shortstack[c]{Pass Rate}
& \shortstack[c]{Latency} \\\midrule
m=\{text\}, g=\{frame\}  & 0.430 & 0.301 & 0.207 & 2.5s \\
m=\{text\}, g=\{clip\}   & 0.433 & 0.293 & 0.157 & 1.5s \\
m=\{vis\}, g=\{frame\}   & 0.477 & 0.336 & 0.323 & 5.2s \\
m=\{vis\}, g=\{clip\}    & 0.507 & 0.369 & 0.293 & 7.3s \\\midrule
m=\{text, vis\}, g=\{frame\}              & 0.513 & 0.352 & 0.303 & 5.0s \\
m=\{text, vis\}, g=\{clip\}               & 0.543 & 0.398 & 0.300 & 5.5s \\
m=\{text\}, g=\{frame, clip\}               & 0.567 & 0.413 & 0.217 & 3.3s \\
m=\{vis\}, g=\{frame, clip\}             & 0.497 & 0.355 & 0.327 & 6.1s \\
\midrule
m=\{text, vis\}, g=\{frame, clip\} 
                          & {0.603} 
                          & {0.433} 
                          & {0.352} 
                          & 4.6s \\ 
\bottomrule
\end{tabular}
\vspace{-0.2cm}
\end{wraptable}
We next ask whether the four configurations synergize or a single one already captures most of the gain, by varying which configurations are available in both stages of \algname{}. In Table \ref{tab:hybrid_comparison}, the top block uses one configuration, the middle block expands along one axis, and the bottom row uses all four. The best single configuration is visual clip-level at $0.507$ Recall@5, the best partial combination is textual frame-and-clip at $0.567$, and the full setting reaches $0.603$ Recall@5, $0.433$ nDCG@5, and $0.352$ pass rate. Combining is not always beneficial, though. Visual frame-and-clip drops to $0.497$ Recall@5, below visual clip-level alone, whereas adding a modality improves over both constituents in both cases. This suggests that cross-modal expansion is the more reliable source of synergy than expanding granularity within a single modality.

Importantly, this synergy comes at a reasonable latency cost. The full setting runs at $4.6$s per query, which is modestly above text-only configurations ($1.5$-$3.3$s) but actually faster than visual-only ones ($5.2$-$7.3$s). This is because each chunk is rendered through its winning configuration, so visual evidence is passed only for chunks that benefit from it, not every retrieved chunk.

\vspace*{-0.1cm}
\subsection{In-depth Analysis of Interleaved Chunk Representations}
\label{sec:result_2_ablation}
\vspace*{-0.1cm}

We further analyze how \algname{}'s interleaved chunk representations work in practice from three angles: the distribution of winning configurations across ranks, the reranking strategy itself, and a comparison against trained query-level routers at the generation stage.

\begin{wraptable}{r}{0.4\textwidth}
\vspace*{-0.3cm}
\caption{Distribution of winning configuration over chunk-aware reranking.}
\vspace*{0.05cm}
\label{tab:rank_memory}
\centering
\scriptsize
\setlength{\tabcolsep}{2.05pt}
\renewcommand{\arraystretch}{1.05}
\begin{tabular}{@{}lccccc@{}}
\toprule
Configuration & Rank 1 & Rank 2 & Rank 3 & Rank 4 & Rank 5 \\
\midrule
\{text, frame\} & 0.37 & 0.27 & 0.25 & 0.27 & 0.25 \\
\{text, clip\} & 0.15 & 0.13 & 0.11 & 0.11 & 0.13 \\
\{vis, frame\} & 0.24 & 0.26 & 0.28 & 0.27 & 0.28 \\
\{vis, clip\} & 0.24 & 0.34 & 0.36 & 0.35 & 0.34 \\
\bottomrule
\vspace*{-0.6cm}
\end{tabular}
\end{wraptable}
\textbf{Distribution of Winning Configurations.} 
Table \ref{tab:rank_memory} shows the distribution of winning configurations across the top-5 ranked chunks. The distribution does not collapse onto a single dominant configuration. At every rank, all four are selected with non-trivial frequency, with $\{$vis, clip$\}$ and $\{$text, frame$\}$ winning most often and $\{$text, clip$\}$ and $\{$vis, frame$\}$ contributing consistently. This spread is not an artifact of score-scale mismatch, as the cross-encoder's scores largely overlap across the four configurations, as detailed in Appendix \ref{app:add_score_calibration}. Different chunks are therefore genuinely best surfaced through different modality--granularity configurations, supporting our design choice of treating both axes as chunk-level decisions.

\begin{wraptable}{R}{0.4\textwidth}
\vspace*{-0.65cm}
\caption{Reranking configuration choice under \algname{}.}
\vspace*{0.1cm}
\label{tab:reranker_config_comparison}
\centering
\scriptsize
\setlength{\tabcolsep}{5.2pt}
\renewcommand{\arraystretch}{1.05}
\begin{tabular}{@{}lccc@{}}
\toprule
{Method} & \shortstack[c]{{R}{@5}} & \shortstack[c]{{nDCG}{@5}} &{{Latency}{}} \\
\midrule
m=\{text\}, g=\{frame\}    & 0.497 & 0.343 & 0.8s \\
m=\{text\}, g=\{clip\}      & 0.490 & 0.339 & 0.5s \\
m=\{vis\}, g=\{frame\}  & 0.547 & 0.391 & 4.1s \\
m=\{vis\}, g=\{clip\}    & 0.560 & 0.396 & 8.0s \\ \midrule
Random         & 0.513 & 0.312 & 3.1s \\
Concatenation & 0.513 & 0.339 & 8.2s \\\midrule
\algname{} (Ours) & {0.603} & {0.433} & 3.4s \\\bottomrule
\end{tabular}\\
\vspace*{-0.4cm}
\end{wraptable}
\textbf{Reranking Configuration Strategy.} 
We next examine whether \algname{}'s reranking rule of rescoring each candidate under its retrieving configuration in Eqs.~\ref{eq:rerank_score}--\ref{eq:winner} is the right strategy. Table~\ref{tab:reranker_config_comparison} compares against three alternatives: (i) applying a single fixed configuration to all candidates regardless of how they were retrieved, (ii) assigning a configuration randomly to each candidate (Random), and (iii) feeding all four configurations jointly into the reranker without selection {as reranked multimodal  late fusion} (Concatenation). \algname{} substantially outperforms all alternatives. This shows that scoring each chunk under its retrieving configuration tag, rather than under a fixed or heuristic alternative, is what drives the gain. In particular, Concatenation fails to match \algname{} despite using all four configurations jointly, indicating that the chunk-level alignment between retrieval and reranking matters more than access to more representations. 

\algname{} also achieves a favorable accuracy--latency trade-off. Reranking visual features is costly due to the larger token count ($4.1$s and $8.0$s for visual-only configurations), but by passing only each chunk's winning configuration, \algname{} runs at $3.4$s while still achieving the best retrieval, substantially faster than visual-only and Concatenation ($8.2$s). The same trend for retrieval performance and latency holds under a different reranker backbone (LamRA-Rank-7B \citep{liu2025lamra}, see Appendix \ref{app:add_reranker_backbone}).

\begin{wraptable}{R}{0.4\textwidth}
\vspace*{-0.7cm}
\caption{Comparison of \algname{} against query-level routing in generation.}
\vspace*{0.1cm}
\label{tab:routing_comparison}
\centering
\scriptsize
\setlength{\tabcolsep}{6.5pt}
\renewcommand{\arraystretch}{1.05}
\begin{tabular}{@{}lcc@{}}
\toprule
Method & {Pass Rate} & {Latency} \\
\midrule
m=\{text\}, g=\{frame\}   & 0.207 & 2.5s \\
m=\{text\}, g=\{clip\}      & 0.157 & 1.5s \\
m=\{vis\}, g=\{frame\}   & 0.323 & 5.2s \\
m=\{vis\}, g=\{clip\}     & 0.293 & 7.3s \\
\midrule
Non-LLM Router w. Training  & 0.329 & 5.7s \\
LLM Router w. Training & 0.310 & 5.3s \\
\midrule
\algname{} (Ours) & \textbf{0.352} & 4.6s \\
\bottomrule
\end{tabular}
\vspace*{-0.4cm}
\end{wraptable}

\textbf{Comparison against Query-Level Routing.}
We also examine whether \algname{}'s chunk-level interleaving of representations is more effective than rendering all retrieved chunks under a single, query-level representation. Table~\ref{tab:routing_comparison} compares \algname{} with (i) fixed-configuration pipelines, which commit to one configuration for every query, and (ii) trained query-level routers, a non-LLM router (MLP) and an LLM router, which choose among all four configurations per query and apply the choice uniformly to all chunks (Appendix~\ref{app:router_implementation}). Given access to every configuration and training on the \dataname{} training split, the routers represent the strongest query-level policy we can construct. Yet \algname{} outperforms every alternative without training. As the routers receive the same retrieved chunks as \algname{}, this margin reflects how evidence is rendered, suggesting that no single configuration suits every chunk. \algname{} sidesteps this by varying the representation chunk by chunk through the same reranking step, at competitive latency.

\vspace*{-0.2cm}
\section{Conclusion}
\vspace*{-0.25cm}

We introduced \texttt{V-RAGBench}, which makes retrieval and generation in \texttt{VideoRAG} separately measurable, and \texttt{CARVE}, which decides modality and granularity per chunk rather than per query. Our results suggest several insights. First, no single configuration suits every chunk, so a chunk-level decision outperforms query-level routing, including trained routers. Second, the decision is most beneficial when carried into generation, where each chunk is rendered under its winning configuration. Third, cross-modal expansion is a more reliable source of synergy than finer granularity within one modality. These findings also hold on third-person videos and a larger configuration space, indicating chunk-level representation selection as a promising direction for \texttt{VideoRAG}.

\subsection*{AI use statement}

In this work, we used generative AI tools for grammar checking and proofreading of author-written text. We have reviewed all AI-assisted edits. Large language and multimodal models are also used as components of the benchmark construction and evaluation pipeline, namely for QA generation, filtering, and LLM-as-a-judge evaluation of \dataname{}, as described in Section~\ref{sec:benchmark} and Appendices~\ref{app:filter_details}, \ref{app:judge}, and \ref{app:prompts}. All references were manually curated and confirmed by the authors.

\subsection*{Ethics statement}

We do not foresee specific ethical risks arising directly from the proposed method or benchmark beyond those generally associated with multimodal retrieval and video understanding research. Nevertheless, we acknowledge that egocentric video inherently captures personal, daily-life activities, and any downstream deployment of \texttt{VideoRAG} systems should consider privacy, consent, and responsible use of personal video data.

For the human evaluation components described in Appendix~\ref{app:human_audit} and Appendix~\ref{app:judge}, we recruited graduate-student annotators to verify the reliability of our filtering pipeline and the LLM-as-a-judge evaluation. All annotators were compensated above the U.S. federal minimum wage, were informed of the task scope and purpose in advance, and participated voluntarily. No personally identifiable information about the annotators is disclosed in the dataset, paper, or supplementary materials, and their contributions are used solely for the validation analyses reported in this work.

\subsection*{Reproducibility statement}

We take several steps to ensure the reproducibility of both \dataname{} and \algname{}. The benchmark construction pipeline is described in Section~\ref{sec:benchmark}, and source video selection, event extraction and deduplication, query generation, and the five post-hoc filters are detailed in Appendices~\ref{app:statistics} to \ref{app:filter_details}, including dataset statistics (Table~\ref{tab:statistics}), domain and category definitions (Tables~\ref{tab:domain_definition} and \ref{tab:query_categories}), and filtering statistics (Table~\ref{tab:filtering_stats}). All prompts used for QA generation, filtering, memory construction, answering, and judging are provided verbatim in Appendix~\ref{app:prompts}, and the human verification protocols for the benchmark and the LLM-as-a-judge are reported in Appendices~\ref{app:human_audit} and \ref{app:judge}. \algname{} is formalized in Section~\ref{sec:method} and Algorithm~\ref{alg:carve}, and its implementation, together with those of all baselines and the trained routers, is specified in Appendix~\ref{app:implimentation}, including model checkpoints (Tables~\ref{tab:open_models} and \ref{tab:closed_models}), hyperparameters, decoding settings, and compute resources. Evaluation metrics are defined in Appendix~\ref{app:metrics}, and run-to-run variance and significance tests are reported in Table~\ref{tab:baseline_comparison} and Appendix~\ref{app:add_stat}. The benchmark, source code, and evaluation scripts will be publicly released upon acceptance.

\bibliographystyle{plainnat}
\bibliography{references}

\clearpage
\clearpage
\newpage
\appendix

\section{Dataset}

\begin{figure}[t]
    \centering
    \includegraphics[width=1\textwidth]{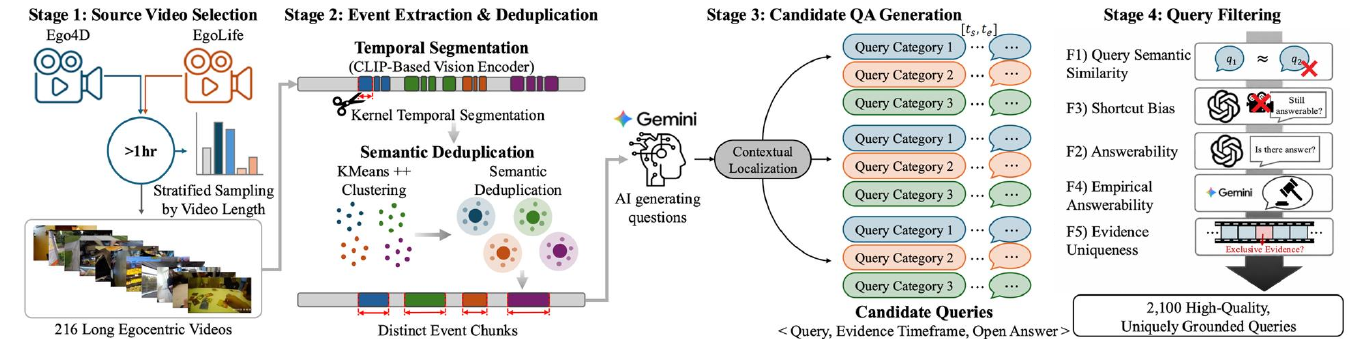}
    \caption{Overview of the V-RAGBench Construction Pipeline. We construct \dataname{} through four stages: source video selection, event extraction and deduplication, candidate QA generation, and query filtering. Hour-scale egocentric videos are segmented into distinct event chunks, used to generate open-ended QA candidates, and then rigorously filtered for redundancy, answerability, shortcut bias, empirical answerability, and evidence uniqueness, yielding a balanced set of uniquely grounded queries.}
    \label{fig:vragbench-pipeline}
\end{figure}

The overall pipeline of \dataname{} construction is described in Figure~\ref{fig:vragbench-pipeline}.

\subsection{Dataset statistics}
\label{app:statistics}
\dataname{} is a benchmark for egocentric video understanding constructed from two complementary sources: long-form daily-life recordings from EgoLife and diverse activity-centric videos from Ego4D. The dataset comprises 2,100 queries grounded in 216 uncut source videos. Detailed dataset statistics, including the distribution of queries across source datasets, categories, domains, and evidence chunk durations, are summarized in Table~\ref{tab:statistics}. 
 
\begin{table}[ht!]
\caption{Dataset statistics detailing the distribution of 2,100 queries across categories, domains, and chunk durations, alongside 216 uncut source videos.}
\footnotesize
\vspace{2pt}
\label{tab:statistics}
\centering
\begin{tabularx}{\columnwidth}{@{}lllX@{}}
\toprule
Scope & Attribute & Sub-category & Count (\% / Details) \\ \midrule
\multirow{3}{*}{Source Videos} 
 & \multicolumn{2}{l}{Total} & 216 \\ \cmidrule(l){2-4}
 & \multirow{2}{*}{Source} & EgoLife & 42 (Mean: 379.1 min, \newline Range: 177.8--552.7 min) \\
 & & Ego4D & 174 (Mean: 86.0 min, \newline Range: 60.0--423.6 min) \\ \midrule
\multirow{16}{*}{Queries} 
 & \multicolumn{2}{l}{Total} & 2,100 \\ \cmidrule(l){2-4}
 & \multirow{2}{*}{Source} & EgoLife & 1,320 (62.86\%) \\
 & & Ego4D & 780 (37.14\%) \\ \cmidrule(l){2-4}
 & \multirow{3}{*}{Category} & Action Flow \& Movement & 700 (600 Train / 100 Test) \\
 & & Interaction & 700 (600 Train / 100 Test) \\
 & & \begin{tabular}[c]{@{}l@{}}Object-Centric \\ Visual Understanding\end{tabular} & 700 (600 Train / 100 Test) \\ \cmidrule(l){2-4}
 & \multirow{6}{*}{Domain} & Daily & 1,320 (62.86\%) \\
 & & Domestic & 203 (9.67\%) \\
 & & Social & 195 (9.29\%) \\
 & & Stationary & 180 (8.57\%) \\
 & & Manual & 169 (8.05\%) \\
 & & Outdoor & 33 (1.57\%) \\ \cmidrule(l){2-4}
 & \multirow{4}{*}{\begin{tabular}[c]{@{}l@{}}Evidence \\ Chunk Duration\end{tabular}} 
 & $<$ 5s & 330 (15.71\%) \\
 & & 5--30s & 1,350 (64.29\%) \\
 & & 30--60s & 168 (8.00\%) \\
 & & 1--5 min & 252 (12.00\%) \\ 
\bottomrule
\end{tabularx}
\end{table}

\subsection{Definitions of the source video domains}\label{app:domain}

Ego4D videos cover heterogeneous activities, while EgoLife videos are uniformly
daily-life. We group Ego4D videos into five high-level domains — \emph{Domestic},
\emph{Social}, \emph{Stationary}, \emph{Manual}, and \emph{Outdoor} — by
collapsing the Ego4D metadata categories. For each in-distribution domain we
sample $36$ training and $6$ test videos based on the within-domain length
distribution; the \emph{Outdoor} domain is held out entirely for test, providing an
out-of-domain split. For EgoLife, one of the seven days per participant is sampled as
test, with non-overlapping day assignments across participants to prevent leakage. The
resulting test set spans both familiar and unseen domains, supporting both standard
and out-of-distribution evaluation. Formal definitions of these domains are provided in Table~\ref{tab:domain_definition}.

\begin{table}[ht!]
\caption{Egocentric video domains covered in \dataname{}}
\label{tab:domain_definition}
\footnotesize
\vspace{2pt}
\centering

\begin{tabularx}{\columnwidth}{@{}p{1.5cm}p{1.8cm}X@{}}
\toprule
Source Dataset & Domain & Definition \\ \midrule
EgoLife & Daily &
  Long-form egocentric daily-life videos from EgoLife, covering extended personal routines across time. Unlike Domestic, which is a specific Ego4D activity domain, Daily refers to broader continuous life-logging episodes that may include household routines, desk work, leisure, transitions, and other ordinary activities across a day. \\ \midrule
\multirow{5}{=}{Ego4D} & Domestic &
  Videos of household or home-centered everyday activities, such as cooking, cleaning, organizing, eating, or moving around indoor living spaces. The key characteristic is routine domestic activity within a home-like environment. \\ \cmidrule(l){2-3}
 & Manual &
  Videos centered on hands-on physical manipulation, craftwork, repair, assembly, tool use, or labor-intensive activities. The key characteristic is that the camera wearer is actively manipulating objects or materials. \\ \cmidrule(l){2-3}
 & Social &
  Videos involving interpersonal interaction, such as conversation, collaboration, instruction, assistance, or shared activity with other people. The key characteristic is that understanding the scene depends on social context, other people's actions, or communicative behavior. \\ \cmidrule(l){2-3}
 & Stationary &
  Low-motion egocentric videos where the camera wearer remains largely fixed in place and attends to a stable target over time, such as a computer screen, television, desk, document, or workspace. The key characteristic is not the presence of a screen itself, but the static first-person viewpoint and sustained attention to the same visual area. \\ \cmidrule(l){2-3}
 & Outdoor &
  Videos captured in outdoor or large-scale physical environments, often involving walking, navigation, travel, sports, outdoor work, or other movement-heavy activities. The key characteristic is that the camera wearer interacts with a changing external environment rather than a fixed indoor workspace or household setting. \\
\bottomrule
\end{tabularx}
\end{table}

\subsection{Definitions of the query categories}\label{app:catagories}

The 2,100 queries are organized into three reasoning categories that target distinct capabilities of egocentric video understanding: \textit{Action Flow \& Movement}, which probes temporal and sequential reasoning; \textit{Object-Centric Visual Understanding}, which focuses on fine-grained recognition of object attributes and states; and \textit{Interaction}, which examines both physical manipulation and social engagement. The precise definitions of each category are given in Table~\ref{tab:query_categories}.

\begin{table}[ht]
\caption{Definitions of the query categories used in the dataset.}
\label{tab:query_categories}
\footnotesize
\vspace{2pt}
\centering
\begin{tabularx}{\columnwidth}{@{}p{3.5cm}X@{}}
\toprule
Category & Description \\ \midrule
Action Flow \newline \& Movement & 
  Questions requiring understanding of event sequences, movement, and scene transitions. These questions emphasizes temporal flow before, during, and after key actions. The expected answer must describe a temporal or sequential fact. \\ \midrule
Object-Centric \newline Visual Understanding & 
  Questions that require recognizing specific details about objects: identity, visual attributes (color, shape, size), current state, or count. The expected answer must identify a specific object detail. \\ \midrule
Interaction & 
  Questions about the camera wearer's interaction with their environment, encompassing both the physical manipulation of objects (how things are held, used, or transferred) and social engagement (roles, relationships, and interpersonal behaviors). The expected answer must describe a physical interaction with an object or a social fact involving other people. \\ 
\bottomrule
\end{tabularx}
\end{table}

\subsection{Event selection and de-duplication details}\label{app:event_stats}

\paragraph{Event Chunk Construction Details.}

Section~\ref{sec:benchmark} describes the overall event extraction and deduplication procedure. Here we clarify two implementation consequences of this design. First, temporal segmentation and semantic deduplication play separate roles: segmentation determines valid time-contiguous event candidates, while deduplication only selects among those candidates and does not modify their temporal boundaries. Thus, each retained chunk remains an original contiguous video interval.

Second, all event-level statistics are computed after deduplication. Since the retained chunks are selected as representatives of semantically similar segment groups, the final evidence pool is intentionally smaller and less redundant than the raw segmentation output. In particular, repeated daily routines are counted only through their selected representative chunks, which reduces the chance that a generated query can be answered from multiple near-duplicate events in the same video.

Because the representatives are not forced into a fixed temporal length, the resulting chunks naturally span a range of durations. We report the duration distribution of these deduplicated event chunks in Table~\ref{tab:event_stats} and Table~\ref{tab:event_distribution}.

\begin{table}[ht]
\centering
\caption{Descriptive statistics of event chunk durations (in seconds) by dataset.}
\label{tab:event_stats}
\vspace{2pt}
\footnotesize
\begin{tabular}{@{}lrrrrr@{}}
\toprule
Split & Count & Mean (s) & Median (s) & Min (s) & Max (s) \\
\midrule
All (Overall) & 19,751 & 28.03 & 8.60 & 4.70 & 300.00 \\
\midrule
EgoLife & 11,109 & 31.65 & 7.60 & 4.70 & 300.00 \\
Ego4D & 8,642 & 23.39 & 11.00 & 5.00 & 300.00 \\
\bottomrule
\end{tabular}
\end{table}

\begin{table}[ht]
\centering
\caption{Distribution of event chunk durations scaled to the final subset of 2,100 queries.}
\vspace{2pt}
\label{tab:event_distribution}
\footnotesize
\begin{tabular}{@{}lrr@{}}
\toprule
Duration Bin (s) & Count & Percentage \\
\midrule
0--5 & 331 & 15.78\% \\
5--10 & 810 & 38.59\% \\
10--30 & 594 & 28.27\% \\
30--60 & 167 & 7.97\% \\
60--120 & 77 & 3.66\% \\
120--300 & 116 & 5.51\% \\
300--500 & 5 & 0.22\% \\
\midrule
Total & 2,100 & 100.00\% \\
\bottomrule
\end{tabular}
\end{table}

\subsection{QA Filter Details}\label{app:filter_details}
Table~\ref{tab:filtering_stats} shows that the filtering pipeline is highly selective: only $5{,}907$ of the initial $67{,}370$ candidates remain after all five stages, corresponding to $8.77\%$ of the original candidate pool. This low retention rate reflects the strict grounding requirements of \dataname{}: retained queries must be semantically non-redundant, answerable from the annotated evidence, robust to shortcut bias, empirically answerable by a strong VLLM, and not recoverable from
non-evidence chunks in the same video. The $5{,}907$ retained queries form the post-filter candidate pool before category balancing. Since the three query categories are unevenly populated after filtering,
we balance the benchmark by downsampling each category to the size of the smallest category, which is \textit{Interaction} (See Table~\ref{tab:query_categories}). We therefore randomly sample $700$ queries per category, yielding the final balanced set of $2{,}100$ queries.

\begin{table}[ht]
\centering
\caption{Filtering statistics for the combined dataset. The table reports the number of queries remaining and the number filtered at each major post-hoc filtering stage. Percentages are relative to the initial candidate count.}
\vspace{4pt}
\label{tab:filtering_stats}
\footnotesize
\begin{tabular}{@{}lrr@{}}
\toprule
Filtering Stage & Remaining & Filtered (\%) \\
\midrule
Initial Candidates & 67,370 & -- \\
Semantic Similarity & 52,783 & 14,587 (21.65\%) \\
Shortcut Bias & 28,331 & 24,452 (36.30\%) \\
Answerability & 19,751 & 8,580 (12.74\%) \\
Empirical Answerability & 11,041 & 8,710 (12.93\%) \\
Evidence Uniqueness & 5,907 & 5,134 (7.62\%) \\
\midrule
Final & 5,907 & (8.77\% kept) \\
\bottomrule
\end{tabular}
\label{tab:filter_detail}
\end{table}

\smallskip
\textbf{Semantic Similarity Filter.}
Event-level visual deduplication does not fully eliminate redundancy at the query level, since different retained events can still induce near-identical natural-language questions. We therefore apply this filter over query embeddings rather than event embeddings, removing candidates that are semantically too close to already retained queries. This reduces benchmark inflation from repeated phrasings of similar actions.

\smallskip
\textbf{Shortcut Bias Filter.}
Some generated queries can be answered from language priors or commonsense knowledge without inspecting the video. To identify such cases, we provide GPT-5.2-chat with the query alone, without any visual context, and compare its answer against the reference answer. If the blind answer matches the reference, the query is discarded as shortcut-biased. Table~\ref{tab:blind_answerable} shows examples of queries filtered as they exhibit shortcut bias.

\smallskip
\textbf{Answerability Filter.}
For each candidate, we evaluate the paired $(q, \text{event clip})$ with GPT-5.2-chat and ask whether the answer can be determined from the visual evidence. Candidates judged unanswerable from their source clip are discarded.

\smallskip
\textbf{Empirical Answerability Filter.}
After the previous filters, we further check whether the query is practically answerable by a strong VLLM using only the source evidence chunk. We run Gemini~2.5-flash on the paired clip and evaluate the generated answer with the binary LLM-as-a-judge described in Appendix~\ref{app:judge}. Queries that the model fails to answer correctly are removed, since they are unlikely to be reliably answerable in downstream evaluation.

\begin{figure}[ht]
    \centering
    \includegraphics[width=0.9\textwidth]{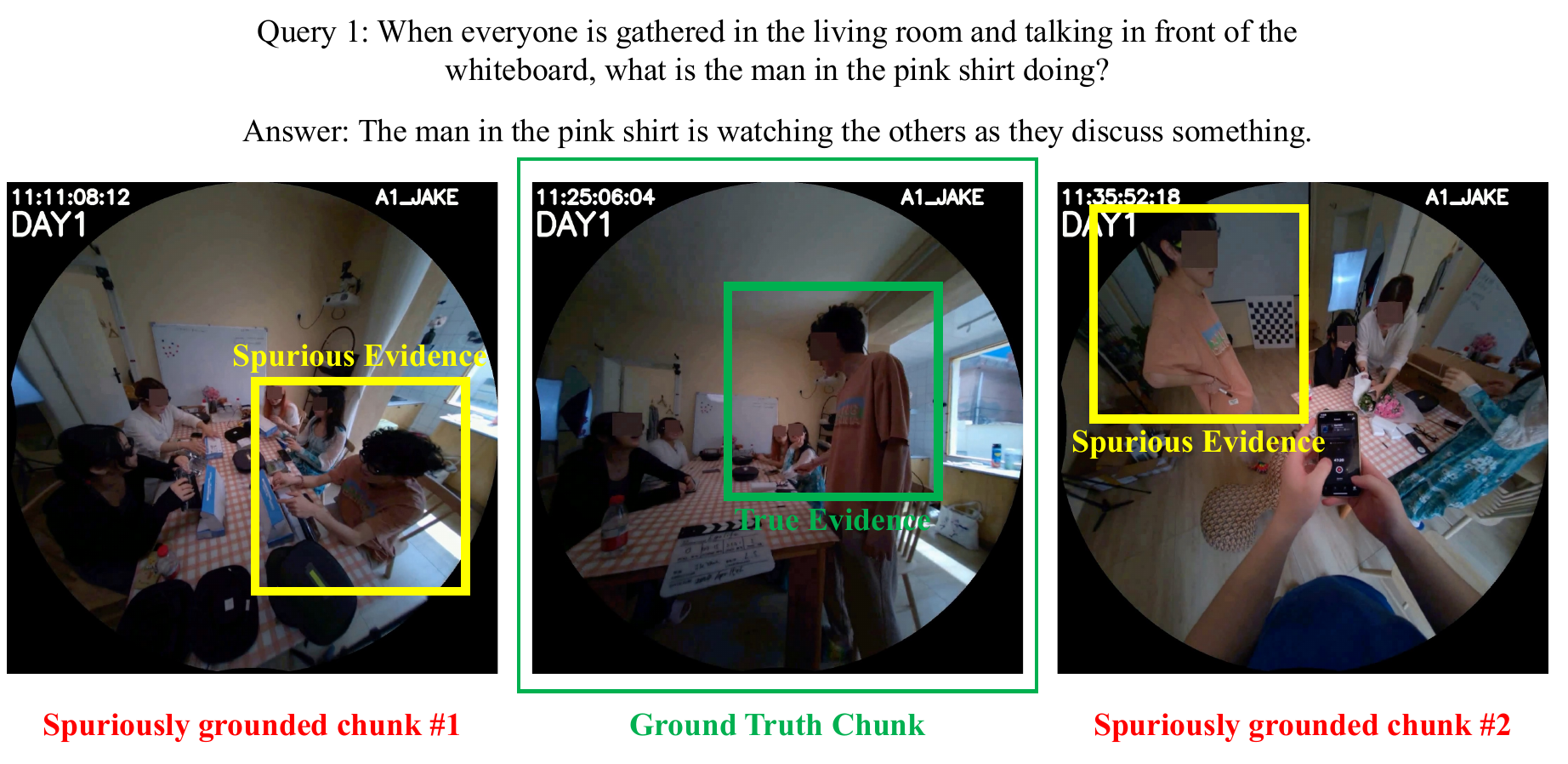}
    \includegraphics[width=0.9\textwidth]{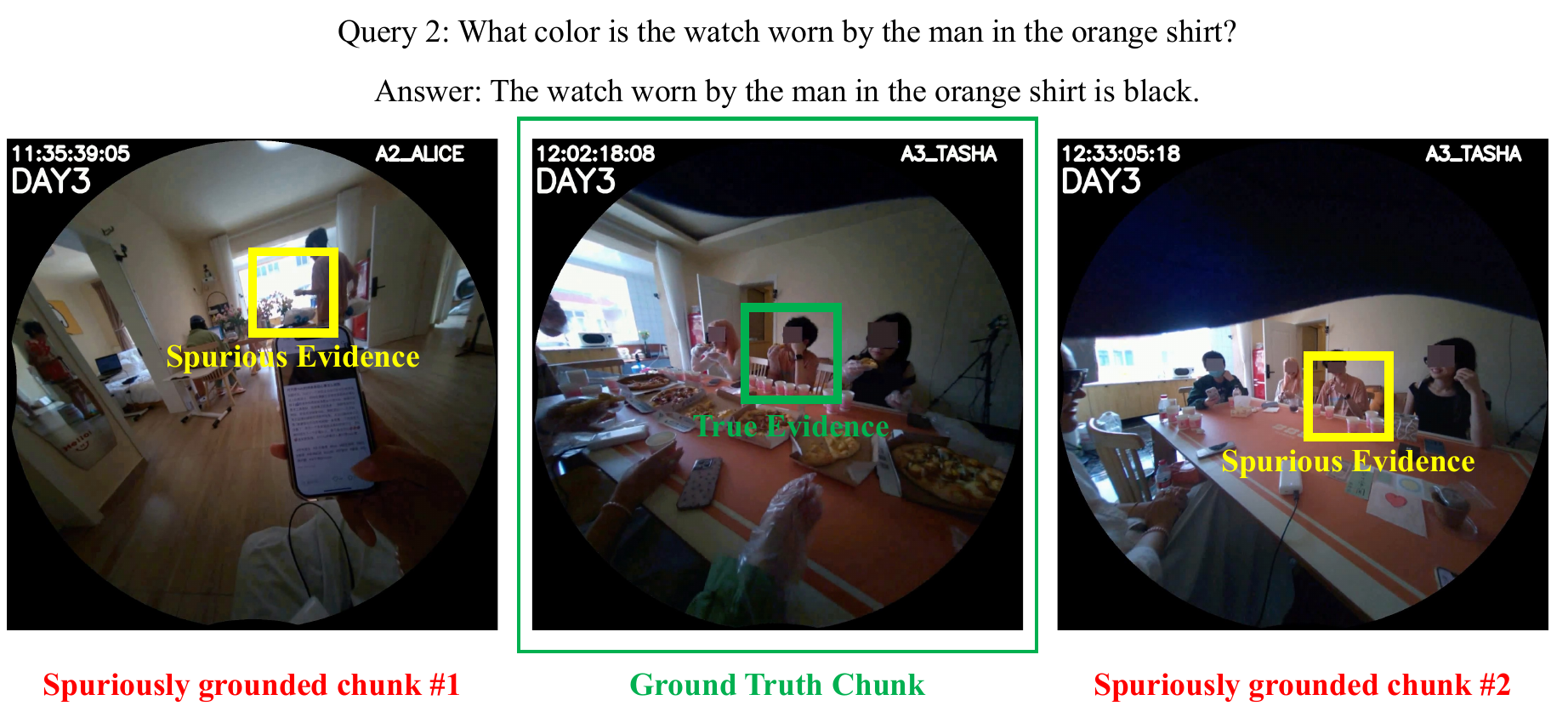}
    \includegraphics[width=0.9\textwidth]{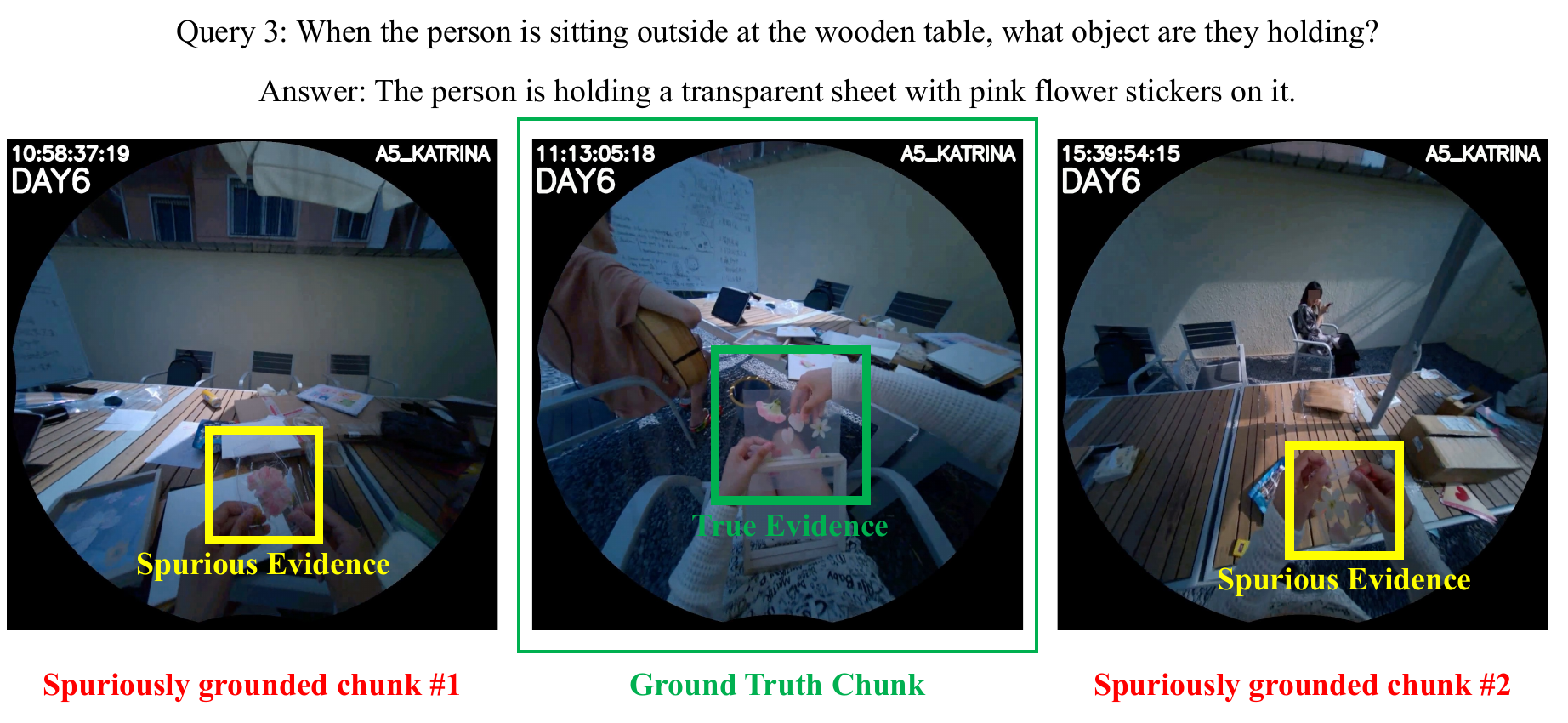}
    \caption{Examples of the queries not grounded with a unique evidence chunk.}
    \label{fig:uniqueness_example}
\end{figure}

\smallskip
\textbf{Evidence Uniqueness Filter.}
To test whether the annotated evidence is uniquely sufficient, we simulate retrieval over non-evidence clips from the same source video. Each video is split into $2$-minute chunks, which are embedded with Qwen3-VL-Embedding-8B. For every query, we retrieve the top-$10$ most similar chunks while explicitly excluding the chunk containing the true evidence. Gemini~2.5-flash is then asked to answer using only this non-evidence retrieval set, and the output is judged with the same binary LLM-as-a-judge as in the empirical answerability filter. If the answer is still judged correct, the query is removed because its answer is recoverable from elsewhere in the video rather than uniquely grounded in the annotated evidence. Figure~\ref{fig:uniqueness_example} presents the case study of queries whose evidence is not uniquely grounded in a single evidence chunk.

\subsection{Human Verification of Query Filters}\label{app:human_audit}

This section details the human verification summarized in Table~\ref{tab:human_audit} of Section~\ref{sec:benchmark}, which consists of three checks, answerability, evidence uniqueness, and query redundancy.

\smallskip
\textbf{Annotators and Protocol.}
Every item in each check is verified independently by two annotators. Annotators are graduate students with at least one first-author publication at a top-tier AI conference and English proficiency equivalent to CEFR C2. Each annotator received a 30-minute instructional briefing before annotation, and for the exhaustive audits additionally completed standardized written training and a calibration session. Annotators label each item without access to the other annotator's judgment, and each check reports the resulting inter-annotator agreement (IAA). We report Cohen's $\kappa$~\citep{cohen1960coefficient} for answerability and Gwet's AC1~\citep{gwet2008computing} for the two exhaustive audits, since $\kappa$ is undefined when no positive case exists and severely understates agreement when positives are around 3\%.

\smallskip
\textbf{Answerability.}
We sample 577 QA pairs at random, corresponding to a 95\% confidence interval with a margin of $\pm 4$ over the full set of 2,100 queries (1,154 judgments). Each annotator is provided with the source video clip, the query, and the ground-truth answer, and judges whether the answer can be reasonably inferred from the video. 96.9\% of the sampled queries are judged answerable by both annotators ($\kappa=0.63$, substantial agreement), confirming that retained queries are reliably answerable from their evidence.

\smallskip
\textbf{Query Redundancy.}
This audit verifies query-level non-redundancy, the property targeted by the Semantic Similarity filter. All 2,648 pairs of test queries drawn from the same video were judged for semantic duplication (5,296 judgments). Neither annotator flagged any pair, with 100\% agreement (AC1=1.0).

\smallskip
\textbf{Evidence Uniqueness.}
This audit verifies that each query is grounded in a single evidence chunk over the entire video, covering properties (i) and (iii) in Section~\ref{sec:benchmark}, and uses no retriever. Annotators watched all 2,094 two-minute windows of the test videos and judged, for each of the 10,206 window--query pairs, whether the query is answerable from that window alone (20,412 judgments). 97.1\% of pairs outside the labeled evidence were judged unanswerable by both annotators (95\% CI 96.8--97.5\%, raw agreement 95.95\%, AC1=0.955), which establishes evidence uniqueness over the full corpus independently of any retriever.

\begin{table}[!t]
\centering
\caption{Examples of benchmark video queries with shortcut bias, which are answerable \emph{without} access to the ground-truth source video based on context and parametric knowledge.}
\vspace{2pt}
\label{tab:blind_answerable}
\footnotesize
\renewcommand{\arraystretch}{1.3}
\begin{tabularx}{\linewidth}{X}
\toprule
\textbf{Q1: While I am standing at the wooden island with the carton of milk and metal mixing bowls, what is the color of the liquid inside the clear plastic measuring cup?} \\
\textit{Blind prediction:} White. \quad \textit{GT:} White. \\
\textit{Why answerable:} The question explicitly mentions a carton of milk on the same surface as the measuring cup. Common sense strongly suggests the liquid being measured is the milk, making ``white'' inferable without watching the video. \\
\midrule
\textbf{Q2: What visible item indicates the woman's professional role while she sits and talks to me?} \\
\textit{Blind prediction:} A visible name badge or ID badge. \quad \textit{GT:} A badge. \\
\textit{Why answerable:} The phrase ``visible item indicates professional role'' is a near-defining description of a name badge or ID badge. Parametric knowledge of workplace conventions makes this the default answer regardless of video content. \\
\midrule
\textbf{Q3: Upon approaching the man sitting at the white table, how does the person acknowledge my presence?} \\
\textit{Blind prediction:} He looks up and greets you (e.g., says hello or nods). \quad \textit{GT:} By greeting. \\
\textit{Why answerable:} Standard social conventions dictate that acknowledging an approaching person involves looking up and greeting them. The query frames the action as ``acknowledge,'' which already implies a greeting-style response. \\
\bottomrule
\end{tabularx}
\end{table}

\section{Implementation details}\label{app:implimentation}

\smallskip
\textbf{Model Specification.} Table~\ref{tab:open_models} lists all open-source models used in our experiments, while Table~\ref{tab:closed_models} summarizes the proprietary models. The total API cost incurred in our experiments is approximately \$1,200, including both QA generation and QA filtering.

\smallskip\textbf{Resources.} All experiments run on $8\times$ NVIDIA H200 ($141$~GB) GPUs with CUDA $13.0$.

\begin{table}[ht!]
\caption{Pretrained models used in our pipeline.}
\label{tab:open_models}
\centering
\footnotesize
\vspace{2pt}
\setlength{\tabcolsep}{4pt}
\renewcommand{\arraystretch}{1.15}
\begin{tabularx}{\columnwidth}{@{}lXX@{}}
\toprule
Model & Checkpoint & Purpose \\
\midrule
Qwen3-VL-8B-Instruct  & Qwen/Qwen3-VL-8B-Instruct      & Video QA, VLM for construct Text Clip and Text Frames in Section~\ref{sec:preliminaries} \\
Qwen3-VL-32B-Instruct   & Qwen/Qwen3-VL-32B-Instruct    & Video QA\\
Gemma-4-26B-A4B  & google/gemma-4-26B-A4B  & Video QA \\
EVA02-E-14-plus & \begin{tabular}[t]{@{}l@{}}timm/eva02\_enormous\_patch14\_plus\_\\clip\_224.laion2b\_s9b\_b144k\end{tabular} & \begin{tabular}[t]{@{}l@{}}Frame embedding extraction for\\segmentation in Section~\ref{sec:benchmark}\end{tabular} \\
Qwen3.6-35B-A3B        & Qwen/Qwen3.6-35B-A3B          & LLM as Judge in Section~\ref{sec:evaluation} \\
Qwen3-VL-Embedding-2B  & Qwen/Qwen3-VL-Embedding-2B    & Embedding model in Section~\ref{sec:method} \\
Qwen3-VL-Reranker-2B   & Qwen/Qwen3-VL-Reranker-2B     & Reranking in Section~\ref{sec:method} \\
LamRA-Rank-7B          & code-kunkun/LamRA-Rank         & Reranking in Section~\ref{sec:method}\\
Qwen3-VL-Embedding-8B  & Qwen/Qwen3-VL-Embedding-8B    & Semantic similarity check in Section~\ref{sec:benchmark} \\
\bottomrule
\end{tabularx}
\end{table}

\begin{table}[ht!]
\caption{Proprietary API used.}
\label{tab:closed_models}
\centering
\footnotesize
\vspace{2pt}
\setlength{\tabcolsep}{4pt}
\renewcommand{\arraystretch}{1.15}
\begin{tabularx}{\columnwidth}{@{}lX@{}}
\toprule
Model & Purpose \\
\midrule
gemini-3-flash-preview   & QA generation in Section~\ref{sec:benchmark}\\
gemini-2.5-flash & Uniqueness, empirical answerability filtering in Section~\ref{sec:benchmark} \\
\midrule
gpt-5.2-chat-latest      & Answerability, Shortcut bias filter in Section~\ref{sec:benchmark}\\
\bottomrule
\end{tabularx}
\end{table}

\subsection{\algname{} Implementations.} 

\smallskip\textbf{Representation Construction.} We split each video into non-overlapping $120$-second segments and build four representations per segment, indexed identically across representations. The VF encodes $5$ keyframes per segment; the VC encodes up to $64$ frames per segment (maximum $1{,}843{,}200$ pixels per frame, sampled at $1$~fps). The TF view captions the $5$ keyframes with Qwen3-VL-8B-Instruct~\citep{qwen3vl2025} up to $512$ tokens; the TC view captions per-second frames at $1$~fps.

\paragraph{Retrieval and Generation.} At inference, we retrieve the top-$K{=}5$ chunks per query, with reranking using the default Qwen3-VL-Reranker-2B~\citep{qwen3vlembedding}. The retrieved chunks are passed to Qwen3-VL-8B-Instruct with a maximum of $512$ new tokens under its default decoding setup. Every generated answer is judged against the reference by a binary judge (Qwen3.6-35B-A3B), using identical judge weights and prompt across our method and all baselines.

\subsection{Baseline Implementations}\label{app:baseline_implementation}
\smallskip
\textbf{System-level Baselines.}\textbf{VideoRAG-A}~\citep{jeong2025videorag} is a retrieval method that encodes each video segment by linearly interpolating its textual and visual representations. We set $\alpha$ as 0.5 in this set up. \textbf{GQR}~\citep{uzan2026guided} refines the retriever's query embedding by gradient descent against a mixture of the two retriever's score distributions, using KL minimization. We set two retrievers as text and video retrievers with learning rate $5\mathrm{e}{-4}$. \textbf{FreeRet}~\citep{zhu2026freeret} repurposes an MLLM (Qwen2-VL-2B-Instruct~\citep{wang2024qwen2} in our setup) as a training-free two-stage retriever, extracting segment embeddings from the last self-attention sublayer and reranking the candidates via MCQ-based scoring. \textbf{GME}~\citep{zhang2025gme} is a unified multimodal embedding model based on Qwen2-VL-2B-Instruct~\citep{wang2024qwen2}. We apply both methods directly to the raw video segments and query.

\smallskip
\textbf{Query-level Baselines.}\textbf{VideoRAG-B}~\citep{luo2025video} prompts Qwen3-VL-8B-Instruct~\citep{qwen3vl2025} to decouple each query into three memory-specific sub-queries, retrieves from each memory, and merges the results before passing them to the generator. Following the original paper, we use one visual retriever and two text retrievers. \textbf{RRF}~\citep{rackauckas2024rag} is the RAG-Fusion variant of reciprocal rank fusion: the MLLM (Qwen3-VL-8B-Instruct~\citep{qwen3vl2025}) rewrites each query into two modality-targeted paraphrases — one for text-clip, one for text-frames — which drive two retrievers operating on the text-clip and text-frame embeddings, respectively. The resulting ranked lists are aggregated by the standard RRF formula with $k{=}60$.
\textbf{DAT}~\citep{hsu2025dat} adaptively balances two retrievers per query using LLM. We set the paper's dense and sparse setting with text clip and frames, $\alpha$ that interpolates the min-max-normalized similarities over the candidate union. \textbf{UniversalRAG}~\citep{yeo2026universalrag} routes each query to a single optimal $(m,g)$ configuration using a trained classifier and applies the selected configuration uniformly to all retrieved chunks. We use LoRA-based router(Qwen3-VL-2B-Instruct) and a encoder-based router (T5Gemma~2 270M~\citep{zhang2025t5gemma} with a linear classification head) and replace the paper's sigmoid-thresholded decoding with argmax at inference for consistency with the top-$K$ retrieval setting used by all other baselines.

\smallskip
\textbf{Router Implementations.}
\label{app:router_implementation}
To evaluate the benefits of chunk-level decisions in generation, we compare \algname{} against two types of query-level routers that select a single configuration $(m, g)$ per query for generation. To train LLM-based Router,  we fine-tune Qwen3-VL-2B-Instruct~\citep{qwen3vl2025} as a multi-label classifier over the four configurations to find best generation configuration. We apply LoRA ($r=32$) and train for 5 epochs with a $2\mathrm{e}{-5}$ learning rate. Labels are derived from our train set, where a configuration is labeled positive if it generates a correct answer. At inference, we select the configuration with the highest predicted score (argmax) and apply it uniformly to all retrieved chunks for generation. To train encoder based router for generation, we train an MLP-based classifier on top of a frozen BGE-large-en-v1.5~\citep{bge_embedding} encoder. We use the frozen \texttt{[CLS]} representation as input to a classification head consisting of three hidden blocks with a size of 512. The model is optimized using binary cross-entropy with a $1\mathrm{e}{-3}$ learning rate. Similar to the LLM router, we use \textit{argmax} at inference to determine a single, query-uniform configuration.

\subsection{Evaluation Metrics}
\label{app:metrics}

We evaluate retrieval and generation separately. Let $\mathcal{Q}$ denote the set of evaluation queries. For each query $q \in \mathcal{Q}$, let $G_q$ be the set of ground-truth evidence chunks annotated for $q$, and let $R_q^K = [r_1, r_2, \ldots, r_K]$ be the ranked list of the top-$K$ retrieved chunks. In our main experiments, we set $K=5$.

\smallskip
\textbf{Recall@K.}
Recall@K measures whether the ground-truth evidence appears in the top-$K$ retrieved chunks. For a query $q$, we define
\begin{equation}
\mathrm{Recall@K}(q)
=
\frac{|R_q^K \cap G_q|}{|G_q|}.
\end{equation}
The dataset-level score is the average over all queries:
\begin{equation}
\mathrm{Recall@K}
=
\frac{1}{|\mathcal{Q}|}
\sum_{q \in \mathcal{Q}}
\mathrm{Recall@K}(q).
\end{equation}
Since each query in \dataname{} is designed to have a uniquely sufficient evidence chunk, this reduces to checking whether that evidence chunk is retrieved within the top-$K$.

\smallskip
\textbf{nDCG@K.}
nDCG@K evaluates not only whether the evidence is retrieved, but also how highly it is ranked. We use binary relevance:
\begin{equation}
\mathrm{rel}_i =
\mathbbm{1}[r_i \in G_q],
\end{equation}
where $r_i$ is the chunk ranked at position $i$. The discounted cumulative gain is
\begin{equation}
\mathrm{DCG@K}(q)
=
\sum_{i=1}^{K}
\frac{\mathrm{rel}_i}{\log_2(i+1)}.
\end{equation}
The ideal discounted cumulative gain is
\begin{equation}
\mathrm{IDCG@K}(q)
=
\sum_{i=1}^{\min(K, |G_q|)}
\frac{1}{\log_2(i+1)}.
\end{equation}
Then,
\begin{equation}
\mathrm{nDCG@K}(q)
=
\frac{\mathrm{DCG@K}(q)}{\mathrm{IDCG@K}(q)}.
\end{equation}
The final nDCG@K is averaged over all queries:
\begin{equation}
\mathrm{nDCG@K}
=
\frac{1}{|\mathcal{Q}|}
\sum_{q \in \mathcal{Q}}
\mathrm{nDCG@K}(q).
\end{equation}

\smallskip
\textbf{Pass Rate.}
For generation, we report aggregate pass rate. Let $\hat{a}_q$ be the model-generated answer for query $q$, and let $a_q$ be the ground-truth answer. We use an LLM-as-a-judge $J(\cdot)$ to determine whether $\hat{a}_q$ is correct:
\begin{equation}
J(q, \hat{a}_q, a_q) \in \{0,1\},
\end{equation}
where $1$ indicates that the generated answer matches the key factual content of the reference answer. The pass rate is
\begin{equation}
\mathrm{PassRate}
=
\frac{1}{|\mathcal{Q}|}
\sum_{q \in \mathcal{Q}}
J(q, \hat{a}_q, a_q).
\end{equation}

\smallskip
\textbf{Latency.}
We additionally report latency for experiments where methods differ in online inference cost. For each query $q$, let $T_q$ denote the wall-clock time required by the evaluated online pipeline. Depending on the experiment, this includes query encoding, retrieval, reranking, evidence rendering, and generation, but excludes offline preprocessing such as video chunking, memory construction and index construction. The average latency is
\begin{equation}
\mathrm{Latency}
=
\frac{1}{|\mathcal{Q}|}
\sum_{q \in \mathcal{Q}}
T_q.
\end{equation}
When reporting retrieval-only latency, $T_q$ includes only the online retrieval and reranking steps; when reporting end-to-end latency, it additionally includes generation.

\subsection{LLM-as-a-Judge Details}\label{app:judge}
For evaluating answer correctness on open-ended questions, we use Qwen3.6-35B-A3B as the backbone model for the LLM-as-a-judge. The prompt used for judgment is shown in Figure~\ref{fig:llmjudge}.

\smallskip
\textbf{Human Verification.} To assess the reliability of the LLM-as-a-judge, we conduct human validation of its outputs at two levels, a random sample over the full set of predictions and an exhaustive audit of the test set. We apply the same annotator qualification criteria and protocol as in Appendix~\ref{app:human_audit}. Each annotator is provided with the model-generated answer, the ground-truth answer, and the binary decision from the LLM-as-a-judge indicating whether the prediction is correct, and independently assesses whether the judge's label is correct.

For the sampled verification, we randomly sample $600$ instances from the full set of $2{,}100$ predictions (corresponding to a $95\%$ confidence interval with a margin of error of $\pm 4$). We find substantial agreement between human annotators and the LLM-as-a-judge, with a Cohen's Kappa score of $0.61$, and both annotators agree that the judge made an incorrect decision in only $2.2\%$ of the instances.

For the exhaustive audit, one evaluation run over the test set yields $300$ binary decisions, and both annotators verified each of them, $600$ judgments in total. $99.0\%$ of the decisions were upheld, and only $3/300$ were unanimously flagged as incorrect. At the level of individual judgments, LLM--human agreement is $95.8\%$ (Cohen's $\kappa$ $0.91$) and human--human agreement $93.7\%$ ($\kappa$ $0.87$), so the judge is as consistent with human annotators as they are with each other. Since the labels are balanced in this audit, $\kappa$ is well-defined, and Gwet's AC1 concurs ($0.88$--$0.92$). These results confirm that the binary LLM-as-a-judge is reliable for evaluating open-ended answers at scale.

\section{\algname{} Algorithm}\label{app:algorithm}
Algorithm~\ref{alg:carve} shows the formalization of the algorithm of \algname{}.

\begin{algorithm}[ht!]
\caption{\algname{}: Chunk-Aware Reranking for Video Evidence}
\label{alg:carve}
{\small
\begin{algorithmic}[1]
\Require
  \Statex \hspace{1.2em} Per-wearer video corpus $\mathcal{V}$ (2-min non-overlapping chunks)
  \Statex \hspace{1.2em} Query $q$
  \Statex \hspace{1.2em} Multimodal encoder $f$ (Qwen3-VL-Embedding-2B by default)
  \Statex \hspace{1.2em} Multimodal cross-encoder $\mathrm{CE}(\cdot,\cdot)$ (Qwen3-VL-Reranker-2B by default)
  \Statex \hspace{1.2em} Captioner MLLM (Qwen3-VL-8B) for textual summaries (cap $512$ tokens)
  \Statex \hspace{1.2em} Keyframe count $n$ (default $5$)
  \Statex \hspace{1.2em} Per-configuration retrieval cap $k_0$, final list size $k$
  \Statex \hspace{1.2em} Generator $\mathcal{G}$
\Ensure
  \Statex \hspace{1.2em} Multi-representation index $\mathcal{I}$
  \Statex \hspace{1.2em} Tagged final top-$k$ list $V_q^{\star} = \{(v,\, (m^{\star}_v, g^{\star}_v))\}$
  \Statex \hspace{1.2em} Predicted answer $\widehat{a}$
\Statex
\Statex \rule{\linewidth}{0.4pt}
\Statex \textsc{Offline: Chunk-wise Multi-Representation Indexing}
\Statex \rule{\linewidth}{0.4pt}
\For{each chunk $v \in \mathcal{V}$}
    \State $\mathcal{K}(v) \gets$ $n$ keyframes from $k$-means++ \emph{centroids} over per-frame embeddings of $v$
    \Statex \hspace{2.6em}\Comment{centroids cover distinct visual states, not uniform time samples}
    \State $\phi_{\mathrm{vis},\mathrm{frame}}(v) \gets \mathrm{MeanPool}\big(\{f(\kappa) : \kappa \in \mathcal{K}(v)\}\big)$
    \State $\phi_{\mathrm{vis},\mathrm{clip}}(v)  \gets f(v)$ \Comment{single embedding over the entire chunk}
    \State $\phi_{\mathrm{text},\mathrm{frame}}(v) \gets f\!\left(\,\Vert_{\kappa \in \mathcal{K}(v)}\,\mathrm{Summarize}(\kappa)\,\right)$
    \Statex \hspace{2.6em}\Comment{per-keyframe summaries concatenated before encoding}
    \State $\phi_{\mathrm{text},\mathrm{clip}}(v)  \gets f\!\left(\mathrm{Summarize}(v)\right)$
    \State $\mathcal{I} \gets \mathcal{I} \cup \big\{\,(m, g, v,\, \phi_{m,g}(v)) : (m,g) \in \{\mathrm{vis},\mathrm{text}\}\times\{\mathrm{frame},\mathrm{clip}\}\,\big\}$
\EndFor
\Statex
\Statex \rule{\linewidth}{0.4pt}
\Statex \textsc{Online: Retrieval, Reranking, and Generation}
\Statex \rule{\linewidth}{0.4pt}
\Statex \textbf{// Stage 1: Parallel candidate pooling --- one query embedding, four index shards}
\State Compute $f(q)$ once
\State $\mathcal{P}(q) \gets \emptyset$ \Comment{configuration-tagged candidate pool}
\For{each $(m,g) \in \{\mathrm{vis},\mathrm{text}\}\times\{\mathrm{frame},\mathrm{clip}\}$}
    \State $C_{(m,g)}(q) \gets \operatorname*{Top\text{-}k_0}_{v \in \mathcal{V}}\, \big\langle f(q),\, \phi_{m,g}(v)\big\rangle$
        \Comment{cf. Eq.~\ref{eq:per_config_retrieval}}
    \State $\mathcal{P}(q) \gets \mathcal{P}(q) \cup \big\{\,(v,\, (m,g)) : v \in C_{(m,g)}(q)\,\big\}$
        \Comment{cf. Eq.~\ref{eq:pool}}
\EndFor
\Statex
\Statex \textbf{// Stage 2: Chunk-adaptive reranking --- score each candidate under the tag that surfaced it}
\For{each $(v,\, (m,g)) \in \mathcal{P}(q)$}
    \State $\tilde{s}\big(q;\, (v, (m,g))\big) \gets \mathrm{CE}\!\left(q,\, \phi_{m,g}(v)\right)$
        \Comment{cf. Eq.~\ref{eq:rerank_score}}
\EndFor
\Statex
\Statex \textbf{// Per-chunk winning configuration: keep each chunk's highest-scoring tag}
\For{each chunk $v$ appearing in $\mathcal{P}(q)$}
    \State $(m^{\star}_v, g^{\star}_v) \gets \displaystyle\argmax_{(m,g)\,:\,(v,(m,g)) \in \mathcal{P}(q)} \tilde{s}\big(q;\, (v, (m,g))\big)$
        \Comment{cf. Eq.~\ref{eq:winner}}
\EndFor
\Statex
\Statex \textbf{// Final cross-modal top-$k$: rank chunks by their winning scores on a comparable scale}
\State $V_q^{\star} \gets \Big\{\,(v,\, (m^{\star}_v, g^{\star}_v)) : v \in \text{top-}k \text{ chunks by } \tilde{s}\big(q;\,(v, (m^{\star}_v, g^{\star}_v))\big)\Big\}$
        \Comment{cf. Eq.~\ref{eq:final_retrieval}}
\Statex
\Statex \textbf{// Modality-interleaved generation: each chunk speaks through its winning representation}
\State $\widehat{a} \gets \mathcal{G}\!\left(q,\; \big\{\,\phi_{m^{\star}_v,\, g^{\star}_v}(v)\,\big\}_{(v,\,(m^{\star}_v, g^{\star}_v))\,\in\, V_q^{\star}}\right)$
        \Comment{cf. Eq.~\ref{eq:gen}}
\State \Return $\mathcal{I},\ V_q^{\star},\ \widehat{a}$
\end{algorithmic}
}
\end{algorithm}

\clearpage

\section{Additional Analysis}\label{app:additional_anal}

This section provides additional analyses that support the main results. We first verify the statistical robustness of the reported gains through repeated runs and an extended query set (Appendix~\ref{app:add_stat}), and show that the advantage of \algname{} does not stem from reranking alone (Appendix~\ref{app:add_baseline_with_reranking}). We then examine generalization to third-person long videos beyond the egocentric domain (Appendix~\ref{app:add_longvideobench}). Next, we analyze failure cases (Appendix~\ref{app:add_error_analysis}) and the robustness of \algname{} to the retrieval depth and the reranker backbone (Appendices~\ref{app:add_topk} and~\ref{app:add_reranker_backbone}). We further verify that reranker scores are comparable across modalities through distribution and normalization analyses (Appendix~\ref{app:add_score_calibration}). Finally, we show that \algname{} scales to an expanded configuration space with ASR and OCR memories (Appendix~\ref{app:add_config_scaling}), and that it can be integrated into agentic frameworks as a retrieval module and compared with a memory-based framework (Appendix~\ref{app:add_agentic}).

\subsection{Statistical Robustness}\label{app:add_stat}

\smallskip
\textbf{Significance Testing.}
Since retrieval is deterministic, we test significance on the generation pass rate. We run the five methods with the highest average pass rate three times per backbone under identical settings and compare \algname{} against the second-best method with a two-sided Welch's $t$-test, which gives $p<0.05$ on every backbone. Table~\ref{tab:run_variance} reports the mean and run-to-run standard deviation, and the means are reported in Table~\ref{tab:baseline_comparison}. The run-to-run standard deviation is smaller than the margin of \algname{} over the runner-up on every backbone. The remaining methods are omitted because their gaps to \algname{} far exceed this variance.

\begin{table}[ht]
\caption{Mean and run-to-run standard deviation of the generation pass rate over three runs. $^{*}p<0.05$ against the second-best method (Welch's $t$-test).}
\label{tab:run_variance}
\footnotesize
\vspace{2pt}
\centering
\setlength{\tabcolsep}{4pt}
\renewcommand{\arraystretch}{1.05}
\begin{tabular}{@{}lccc@{}}
\toprule
Method & Qwen3-VL-8B & Qwen3-VL-32B & Gemma-4-26B \\
\midrule
VideoRAG-A   & 0.2378 $\pm$ 0.0201 & 0.3123 $\pm$ 0.0040 & 0.3045 $\pm$ 0.0021 \\
VideoRAG-B   & 0.3117 $\pm$ 0.0136 & 0.3034 $\pm$ 0.0204 & 0.2355 $\pm$ 0.0051 \\
UniversalRAG & 0.2588 $\pm$ 0.0297 & 0.2534 $\pm$ 0.0115 & 0.2279 $\pm$ 0.0252 \\
GME          & 0.2533 $\pm$ 0.0115 & 0.2823 $\pm$ 0.0139 & 0.2116 $\pm$ 0.0044 \\
\midrule
\textbf{\algname{}} & \textbf{0.3523} $\pm$ 0.0108$^{*}$ & \textbf{0.3680} $\pm$ 0.0010$^{*}$ & \textbf{0.3180} $\pm$ 0.0051$^{*}$ \\
\bottomrule
\end{tabular}
\end{table}

\smallskip
\textbf{Consistency on an Extended Query Set.}
The official test split is fixed to preserve its balance across the three query categories and the held-out Outdoor domain (Appendix~\ref{app:domain}). To confirm that method rankings do no on the particular queries in this split, we add 200 randomly drawn queries that were not used to train any method and evaluate all methods on the resulting 500 queries. The pass-rate rankings are virtually identical between the two sets. Spearman $\rho$ is $1.000$ on Qwen3-VL-8B, a perfect match, and $0.988$ on Qwen3-VL-32B and Gemma-4-26B, where only one of 45 method pairs is reversed. \algname{} ranks first on all three backbones in both sets, so the reported gains reflect the methods rather than the sampled queries.

\subsection{Baseline with Reranking}\label{app:add_baseline_with_reranking}
\begin{table}[ht]
\caption{Baselines with reranking. Best values in bold.}
\label{tab:retrieval_comparison}
\footnotesize
\vspace{2pt}
\centering
\setlength{\tabcolsep}{4pt}
\renewcommand{\arraystretch}{1.05}
\begin{tabular}{@{}lcc@{}}
\toprule
Method & Recall@5 & nDCG@5 \\
\midrule
 VideoRAG-A         & 0.510 & 0.343 \\
 VideoRAG-B         & 0.487 & 0.330 \\
GQR                 & 0.503 & 0.341 \\
RRF                 & 0.463 & 0.324 \\
UniversalRAG        & 0.447 & 0.324 \\
UniversalRAG-LORA   & 0.470 & 0.344 \\
DAT                 & 0.460 & 0.315 \\
FreeRet             & 0.263 & 0.212 \\
GME                 & 0.413 & 0.305 \\
\midrule
\textbf{\algname{}} & \textbf{0.603} & \textbf{0.433} \\
\bottomrule
\end{tabular}
\end{table}

To verify that \algname{}'s gains are not solely attributable to the reranking stage, we apply all baselines with the same reranker used in our method. As illustrated in Table~\ref{tab:retrieval_comparison}, while the addition of the reranker generally yields performance improvements, these gains remain marginal and, \algname{} significantly outperforms even the reranked baseline. This result demonstrates that the advantage of \algname{} does not merely stem from the reranking mechanism as a standalone tool. Instead, it originates from the system’s unique ability to select and interleave the most discriminative modality-granularity configurations at the chunk level.

\subsection{Generalization to Third-Person Long Videos}\label{app:add_longvideobench}

\begin{table}[ht]
\caption{Retrieval and generation performance on LongVideoBench validation set. Best values in bold.}
\footnotesize
\vspace{2pt}
\label{tab:longvideobench}
\centering
\setlength{\tabcolsep}{3pt}
\renewcommand{\arraystretch}{1.0}
\begin{tabular}{@{}lcccc@{}}
\toprule
 & Retrieval & \multicolumn{3}{c}{Generation} \\
\cmidrule(lr){2-2} \cmidrule(lr){3-5}
Method & nDCG@5 & Qwen3-VL-8B & Qwen3-VL-32B & Gemma-4-26B \\
\midrule
VideoRAG-A        & 0.504 & 0.631 & 0.709 & 0.631 \\
GQR               & 0.466 & 0.589 & 0.567 & 0.518 \\
FreeRet           & 0.360 & 0.638 & 0.674 & 0.539 \\
GME               & 0.442 & 0.652 & 0.645 & 0.567 \\
RRF               & 0.393 & 0.596 & 0.624 & 0.560 \\
DAT               & 0.440 & 0.617 & 0.631 & 0.610 \\
VideoRAG-B        & 0.494 & 0.660 & 0.674 & 0.624 \\
UniversalRAG      & 0.496 & 0.667 & 0.709 & 0.638 \\
UniversalRAG-LoRA & 0.481 & 0.638 & 0.660 & 0.631 \\
\midrule
\textbf{\algname{}} & \textbf{0.545} & \textbf{0.695} & \textbf{0.766} & \textbf{0.645} \\
\bottomrule
\end{tabular}
\end{table}

We further evaluate whether \algname{} generalizes beyond the challenging egocentric videos. Since \algname{} selects the configuration of each chunk based on its content, the selected configurations can naturally vary across different video domains. We evaluate \algname{} against eight baselines on the validation set of LongVideoBench~\citep{wu2024longvideobench}, which contains third-person long videos spanning ten categories, including lectures, movie recaps, and news. To provide a sufficiently challenging retrieval setting, we restrict the evaluation to the longest duration group, (900, 3600] seconds, resulting in 141 questions. Because LongVideoBench does not guarantee a unique supporting chunk for each query, we report nDCG@5 as a reference metric and focus primarily on generation accuracy. As shown in Table~\ref{tab:longvideobench}, \algname{} achieves the highest nDCG@5 and the best generation accuracy across all three generators. These results show that the benefits of chunk-level configuration selection extend beyond egocentric videos to third-person long videos across diverse domains.

\subsection{Error Analysis on Retrieval and Generation}\label{app:add_error_analysis}

\begin{table}[ht!]
\caption{Retrieval cases where generation fails, across two generator scales.}
\label{tab:failure_cases}
\footnotesize
\vspace{2pt}
\centering
\setlength{\tabcolsep}{4pt}
\renewcommand{\arraystretch}{1.05}
\begin{tabular}{@{}lccc@{}}
\toprule
Case & Retrieval & Qwen3-VL-8B & Qwen3-VL-32B \\
\midrule
Case 1 & Fail & 89  & 86  \\
Case 2 & Hit  & 105 & 104 \\
\bottomrule
\end{tabular}
\end{table}

\begin{table}[ht!]
\caption{Representative failure cases of \algname{} on V-RAGBench.}
\label{tab:failure_examples}
\vspace{2pt}
\footnotesize
\centering
\setlength{\tabcolsep}{5pt}
\renewcommand{\arraystretch}{1.25}
\begin{tabular}{@{}p{0.14\linewidth}p{0.395\linewidth}p{0.395\linewidth}@{}}
\toprule
 & Case 1. Retrieval failure & Case 2. Generation failure \\
\midrule
Query & After I pass the red refrigerator in the hallway, what room do I enter next? & After I place a rolled tortilla on the black stovetop, what is the next action I perform? \\
\addlinespace[2pt]
Retrieval & Fail. The ground-truth chunk is not retrieved. & Hit. The ground-truth chunk is retrieved. \\
\addlinespace[2pt]
Ground truth & A room with a whiteboard and a projector screen. & Picking up a fresh tortilla. \\
\addlinespace[2pt]
Generated answer & \textit{``After passing the red refrigerator in the hallway, you enter the dining area \ldots\ which contains a large table covered with a checkered tablecloth, flowers, and groceries.''} & \textit{``The next action you perform to continue the assembly is adding more filling to the tortilla. \ldots\ after placing the first rolled tortilla on the griddle, you immediately pick up another tortilla, scoop filling from the pan, and roll it.''} \\
\midrule
Cause & The query anchor recurs throughout the video, so a visually similar but incorrect occurrence is retrieved. & The correct action appears in the reasoning, but the generator fails to order the actions in time. \\
\bottomrule
\end{tabular}
\end{table}

To analyze the distribution of failures, we categorize the failed queries into two types under the Qwen3-VL-8B and Qwen3-VL-32B generators. In \textit{Case 1}, both retrieval and generation fail. In \textit{Case 2}, retrieval succeeds but the generator fails to produce the correct answer. As shown in Table~\ref{tab:failure_cases}, both failure types occur at comparable rates across the two generator scales, with Case 2 slightly more frequent than Case 1 for both the 8B and 32B generators.

\smallskip
\textbf{Qualitative analysis.}To understand the source of each failure type, we examine a representative example for each case in Table~\ref{tab:failure_examples}. In Case 1, the key anchor in the query, the red refrigerator, appears multiple times throughout the video. The reranker assigns high scores to these similar occurrences, causing the ground-truth chunk to be missed during retrieval. This highlights a key challenge in long-video retrieval. In Case 2, the correct chunk is successfully retrieved, yet the model still produces an incorrect answer. The generated response correctly identifies the next action, picking up another tortilla, while describing the evidence, but ultimately answers with a later action, adding more filling. The generator therefore identifies the relevant evidence but fails to preserve the temporal order of the actions, indicating a temporal reasoning failure rather than a retrieval or recognition failure.

\subsection{Effect of Top-K}\label{app:add_topk}

We also vary the final retrieval depth $k$ to test whether \algname{} is sensitive to the number of chunks passed to the generator. This setting is important because increasing $k$ can improve recall by adding more candidate evidence, but it can also introduce distractor chunks that make generation harder. A robust \texttt{VideoRAG} method should therefore maintain strong performance across different retrieval depths, rather than relying on a single tuned value of $k$.

Across the tested values of $k$, \algname{} remains stable and consistently competitive, showing that its gains do not depend on a particular retrieval budget. This indicates that chunk-wise parallel reranking improves the ordering of evidence itself, rather than merely benefiting from retrieving more chunks. In other words, \algname{} is effective because it selects better-ranked evidence, not because it relies on a larger context window. This experiment shows that \textbf{chunk-adaptive reranking produces a robust evidence ranking across retrieval depths, making \algname{} less sensitive to the final top-$k$ choice}.

\begin{figure}[ht!]
    \centering
    \includegraphics[width=1.0\linewidth]{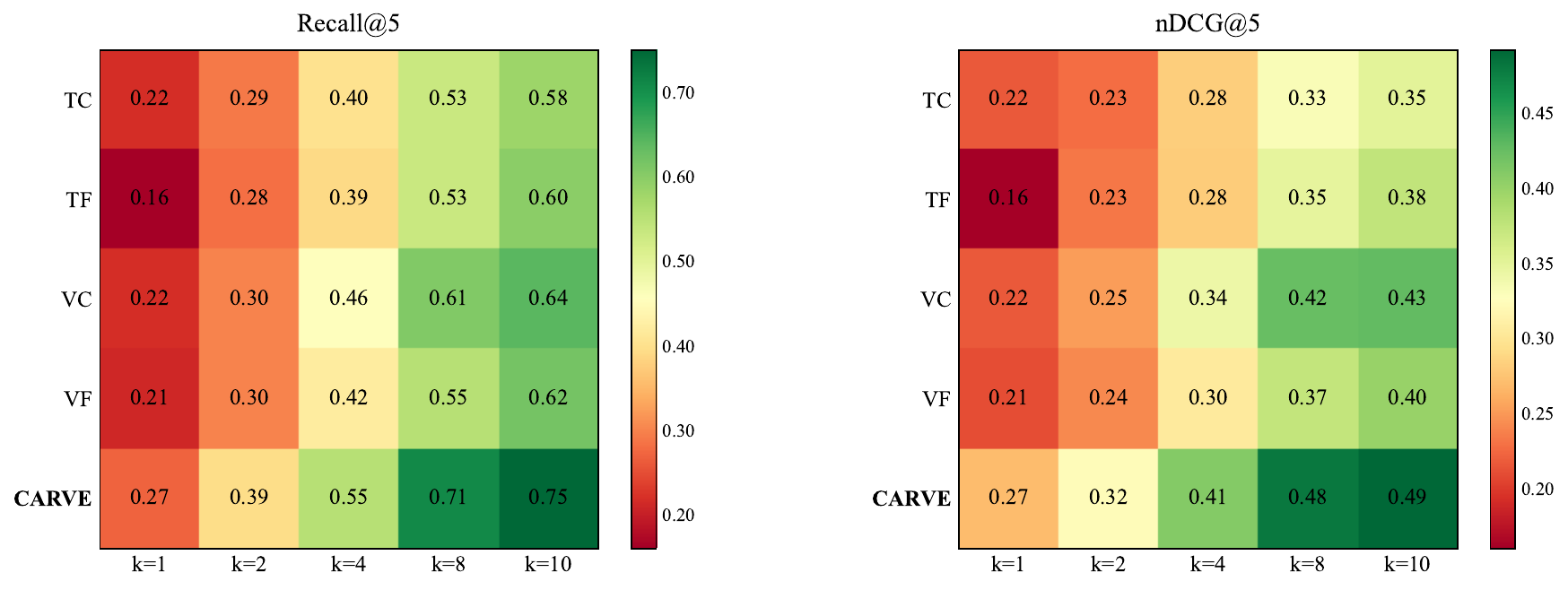}
    \caption{Heatmaps of Recall@k and nDCG@k under different final retrieval depths.}
    \label{fig:topk_heatmap}
\end{figure}

\subsection{Reranker Backbone Robustness}\label{app:add_reranker_backbone}

\begin{table}[ht]
\caption{Reranking configuration choice under \algname{} with LamRA-Rank-7B. Best retrieval values in bold.}
\footnotesize
\vspace{2pt}
\label{tab:reranker_comparison}
\centering
\setlength{\tabcolsep}{3pt}
\renewcommand{\arraystretch}{1.0}
\begin{tabular}{@{}lccc@{}}
\toprule
Method & R@5 & nDCG@5 & Latency \\\midrule
m=\{text\}, g=\{frame\} & 0.520 & 0.339 & 1.0s \\
m=\{text\}, g=\{clip\}  & 0.503 & 0.318 & 0.5s \\
m=\{vis\}, g=\{frame\}  & 0.537 & 0.357 & 4.5s \\
m=\{vis\}, g=\{clip\}   & 0.563 & 0.380 & 8.7s \\
\midrule
Random                  & 0.537 & 0.347 & 4.4s \\
Concatenation           & 0.543 & 0.367 & 11.3s \\
\midrule
\textbf{\algname{}}     & \textbf{0.577} & \textbf{0.401} & 4.0s \\
\bottomrule
\end{tabular}
\end{table}

Table~\ref{tab:reranker_comparison} shows the retrieval performance of each representation configuration under a different reranker backbone, LamRA-Rank-7B. As with Qwen3-VL-Reranker-2B, \algname{} achieves the best R@5 and nDCG@5, outperforming both the strongest single configuration and the Concatenation variant while requiring less latency than Concatenation. Since LamRA-Rank-7B differs from Qwen3-VL-Reranker-2B in both model architecture and prompt template, this confirms that the advantage of chunk-level configuration selection does not depend on a particular reranker or prompt format.

\subsection{Score Calibration Analysis}\label{app:add_score_calibration}

\begin{table}[ht]
\caption{Wasserstein-1 distance between reranker score distributions of different configurations.}
\small
\vspace{2pt}
\label{tab:score_w1}
\centering
\setlength{\tabcolsep}{3pt}
\renewcommand{\arraystretch}{1.0}
\begin{tabular}{@{}llc@{}}
\toprule
Pair & Type & $W_1$ \\
\midrule
Text (all) vs. Visual (all)     & Modality       & 0.074 \\
\midrule
Text clip vs. Text frame        & Within text    & 0.066 \\
Visual clip vs. Visual frame    & Within visual  & 0.004 \\
\midrule
Text frame vs. Visual frame     & Cross-modality & 0.053 \\
Text frame vs. Visual clip      & Cross-modality & 0.054 \\
Text clip vs. Visual clip       & Cross-modality & 0.099 \\
Text clip vs. Visual frame      & Cross-modality & 0.098 \\
\bottomrule
\end{tabular}
\end{table}

\begin{table}[ht]
\caption{Effect of reranking score normalization. Best values in bold.}
\vspace{2pt}
\small
\label{tab:score_norm}
\centering
\setlength{\tabcolsep}{3pt}
\renewcommand{\arraystretch}{1.0}
\begin{tabular}{@{}lcc@{}}
\toprule
Method & R@5 & nDCG@5 \\
\midrule
Z-score (modality)     & 0.580 & 0.354 \\
Min-max (modality)     & 0.583 & 0.347 \\
Z-score (configuration) & 0.547 & 0.347 \\
\midrule
\textbf{Raw score (\algname{})} & \textbf{0.603} & \textbf{0.433} \\
\bottomrule
\end{tabular}
\end{table}

In \algname{}, candidates retrieved under textual and visual configurations are ranked together by their raw cross-encoder scores. This design assumes that the scores are comparable across modalities, since a systematic bias toward one modality would make the final ranking reflect the modality rather than evidence relevance. We verify this assumption from three perspectives. All scores lie on the $[0,1]$ scale of Qwen3-VL-Reranker-2B.

\smallskip
\textbf{Overall score distributions.}
Table~\ref{tab:score_w1} reports the Wasserstein-1 distance ($W_1$) between score distributions. The distance between textual and visual candidates is only 0.074, and every pairwise distance among the four configurations stays below 0.1. Moreover, the distance within the textual modality is 0.066, which is nearly as large as the cross-modal distance, whereas two cross-modality pairs are even smaller at 0.053 and 0.054. This shows that score differences vary across individual configurations rather than modalities, indicating no bias toward either modality.

\smallskip
\textbf{Scores in the top-ranked region.}
We further examine the high-score region, where candidates from different modalities directly compete for the final top-$k$. In this region, the textual and visual distributions nearly coincide, with 15.2\% of textual candidates and 14.9\% of visual candidates scoring 0.6 or higher. The upper percentiles also show no consistent direction, as the 75th percentile is 0.527 for text and 0.555 for visual, whereas the 90th percentile is 0.652 for text and 0.629 for visual. The minor distributional gap in Table~\ref{tab:score_w1} therefore arises from the low score range, which rarely affects the final top-$k$ ranking. This is consistent with Table~\ref{tab:rank_memory}, where all four configurations are regularly selected at every rank, with no modality dominating.

\smallskip
\textbf{Score normalization.}
To examine whether the cross-modal reranking in \algname{} is biased toward a particular modality, we compare it with variants that remove scale differences through normalization. For each query, we normalize the reranker scores in the candidate pool using per-modality z-score, per-modality min-max, or per-configuration z-score normalization, and then reselect the final top-$k$ from the same pool based on the normalized scores. As shown in Table~\ref{tab:score_norm}, all three methods degrade retrieval. This indicates that the raw score differences across configurations are not a modality bias but a relevance signal, since a configuration that reveals a chunk well assigns it a higher score.

\subsection{Scalability to an Expanded Configuration Space}\label{app:add_config_scaling}

\begin{table}[ht]
\caption{Accuracy on Video-MMMU with six configurations. Best values in bold.}
\footnotesize
\vspace{2pt}
\label{tab:config_scaling_perf}
\centering
\setlength{\tabcolsep}{3pt}
\renewcommand{\arraystretch}{1.0}
\begin{tabular}{@{}lccc@{}}
\toprule
Method & Qwen3-VL-8B & Qwen3-VL-32B & Gemma-4-26B \\
\midrule
VideoRAG-A       & 0.705 & 0.773 & 0.761 \\
VideoRAG-B   & 0.716 & 0.750 & 0.716 \\
UniversalRAG-LLM& 0.682 & 0.705 & 0.727 \\
DAT         & 0.705 & 0.727 & 0.727 \\
RRF   & 0.659 & 0.739 & 0.716 \\
\midrule
\textbf{\algname{}} & \textbf{0.750} & \textbf{0.784} & \textbf{0.773} \\
\bottomrule
\end{tabular}
\end{table}

\begin{table}[ht]
\caption{Candidate pool growth and online latency as the configuration space expands on Video-MMMU.}
\footnotesize
\vspace{2pt}
\label{tab:config_scaling_cost}
\centering
\setlength{\tabcolsep}{3pt}
\renewcommand{\arraystretch}{1.0}
\begin{tabular}{@{}clccccc@{}}
\toprule
\#Config & Configurations & Raw cand. & Unique cand. & Dup. rate & Marginal gain & Latency (ms) \\
\midrule
1 & VC                   & 3  & 3.00 & 0.0\%  & --   & 254.7 \\
2 & VC + VF              & 6  & 4.26 & 29.0\% & 1.26 & 346.0 \\
4 & VC + VF + TC + TF    & 12 & 5.72 & 52.3\% & 0.73 & 416.3 \\
6 & 4 Configs + ASR + OCR & 18 & 6.75 & 62.5\% & 0.51 & 528.9 \\
\bottomrule
\end{tabular}
\end{table}

\smallskip
\textbf{Effectiveness with additional configurations.} Since parallel candidate pooling and chunk-adaptive reranking are agnostic to the number of configurations, \algname{} admits new modality-specific memories without architectural changes. We verify this on Video-MMMU~\citep{hu2026video}, a benchmark of lecture videos in which spoken narration and on-screen text carry essential information. We include videos longer than ten minutes and questions that require audio or on-screen text, resulting in 88 questions in total. We then extend the configuration space from four to six by adding an ASR memory transcribed with whisper-large-v3-turbo~\citep{radford2023robust} and an OCR memory extracted with PP-OCRv5~\citep{cui2026pp}. We compare against baselines that can incorporate additional memory types and exclude methods whose memory space is fixed by design. Since Video-MMMU provides no evidence-interval annotations, we report only end-to-end accuracy. As shown in Table~\ref{tab:config_scaling_perf}, \algname{} achieves the best accuracy with all three generators. Notably, \algname{} consistently outperforms VideoRAG-A despite using only the best configuration for each retrieved chunk, while VideoRAG-A provides the generator with all available memory representations. This suggests that selecting the right representation for each chunk is more effective than simply providing the generator with more information.

\smallskip\textbf{Computational overhead.}
Table~\ref{tab:config_scaling_cost} reports how the candidate pool and online latency scale as configurations are added under the same setting. The number of raw candidates grows linearly from 3 to 18, whereas the number of unique candidates grows only $2.25\times$ from 3.00 to 6.75. This gap arises because different modalities and granularities often retrieve the same chunk, which raises the duplication rate to 62.5\% and reduces the marginal gain from 1.26 to 0.51 unique chunks per added configuration. Consequently, total online latency increases by only $2.08\times$, from 254.7\,ms to 528.9\,ms. The additional memory overhead is also negligible, as incorporating ASR and OCR increases the total memory footprint by only 0.033\%.

\subsection{Compatibility with Agentic VideoRAG}\label{app:add_agentic}

\begin{table}[ht]
\caption{Compatibility of \algname{} with agentic VideoRAG frameworks.}
\small
\vspace{2pt}
\label{tab:agentic}
\centering
\setlength{\tabcolsep}{4pt}
\renewcommand{\arraystretch}{1.0}
\begin{tabular}{@{}lcc@{}}
\toprule
Method & nDCG@5 & Pass Rate \\
\midrule
WorldMM                 & 0.292          & 0.133          \\
WorldMM + \algname{}    & \textbf{0.427} & \textbf{0.360} \\
\midrule
VideoAgent              & 0.210          & 0.193          \\
VideoAgent + \algname{} & \textbf{0.281} & \textbf{0.273} \\
\bottomrule
\end{tabular}
\end{table}

\begin{table}[ht]
\caption{Comparison with EgoRAG. Best values in bold.}
\small
\vspace{2pt}
\label{tab:egorag}
\centering
\setlength{\tabcolsep}{3pt}
\renewcommand{\arraystretch}{1.0}
\begin{tabular}{@{}lccccc@{}}
\toprule
 & \multicolumn{2}{c}{Retrieval} & \multicolumn{3}{c}{Generation} \\
\cmidrule(lr){2-3} \cmidrule(lr){4-6}
Method & R@5 & nDCG@5 & Qwen3-VL-8B & Qwen3-VL-32B & Gemma-4-26B \\
\midrule
EgoRAG              & 0.283 & 0.191 & 0.193 & 0.203 & 0.137 \\
\midrule
\textbf{\algname{}} & \textbf{0.603} & \textbf{0.433} & \textbf{0.352} & \textbf{0.368} & \textbf{0.318} \\
\bottomrule
\end{tabular}
\end{table}

Recent VideoRAG research has moved toward agentic frameworks that search for evidence through iterative reasoning or structured memory. To examine whether \algname{} is complementary to these frameworks, we plug \algname{} into agentic methods as their retrieval module. We additionally compare \algname{} with a framework whose retrieval depends on its memory structure.

\smallskip
\textbf{Integration into agentic frameworks.}
Agentic video retrieval frameworks use multi-step reasoning to determine what information to retrieve and whether additional retrieval is needed. However, within each retrieval step, they typically represent all video chunks using a fixed configuration. \algname{} complements this process by selecting the most suitable representation for each chunk, without modifying the agent's planning. We therefore integrate \algname{} into two representative frameworks, WorldMM~\citep{yeo2025worldmm} and VideoAgent~\citep{wang2024videoagent}, replacing only their retrieval modules while keeping all other components unchanged. As shown in Table~\ref{tab:agentic}, this simple integration improves both retrieval and generation performance for both frameworks. The pass rate increases from 0.133 to 0.360 for WorldMM and from 0.193 to 0.273 for VideoAgent. These results show that chunk-level representation selection remains effective in agentic retrieval settings and can be directly incorporated into existing frameworks.

\smallskip
\textbf{Comparison with hierarchical memory retrieval.}
EgoRAG~\citep{yang2025egolife} organizes long videos into a temporal hierarchy of memories and retrieves evidence by progressively narrowing the search from coarse to fine temporal scales. Each query thus follows a single coarse-to-fine search path before the final clips are retrieved. In contrast, \algname{} operates at the chunk level and selects the most suitable representation for each chunk independently. Because EgoRAG's retrieval is tightly coupled with its hierarchical memory structure, we compare against the full EgoRAG pipeline rather than replacing its retrieval module with \algname{}. As shown in Table~\ref{tab:egorag}, \algname{} improves R@5 from 0.283 to 0.603 and increase the pass rate across all three generators. In EgoRAG, early retrieval errors at coarse levels can propagate to later stages, whereas \algname{} scores the original chunks directly without sequentially restricting the search space.

\section{Prompts}\label{app:prompts}
\subsection{QA Generation Prompt}

\begin{figure}[ht]
    \centering
    \includegraphics[width=0.8\textwidth]{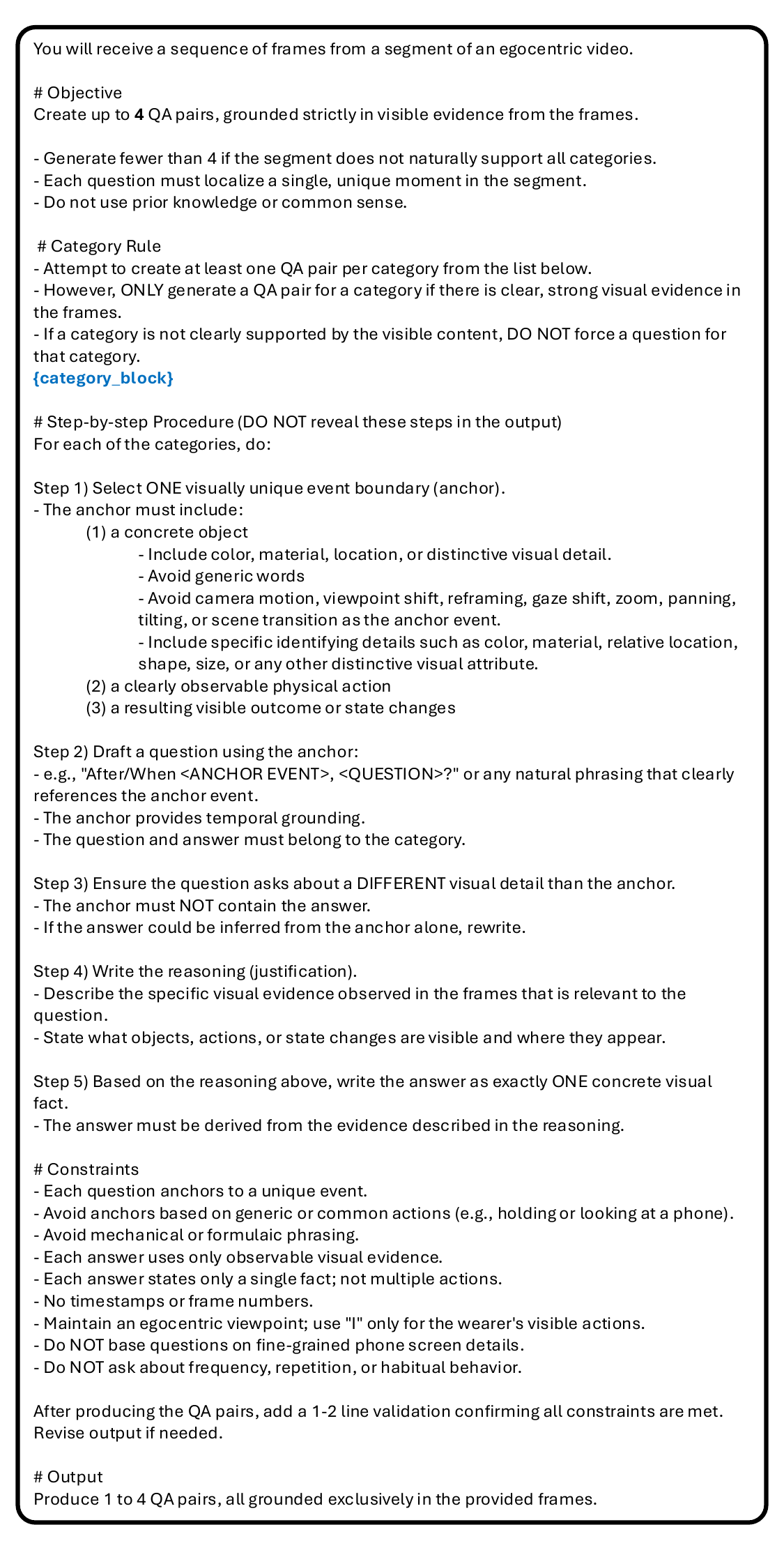}
    \caption{Prompt for QA Generation.}
    \label{fig:qageneration_qageneration}
\end{figure}

Figure~\ref{fig:qageneration_qageneration} presents the prompt used to generate visually grounded QA pairs from egocentric video segments, with each question anchored to a unique observable event in the frames.

\subsection{QA Filtering Prompt}

\begin{figure}[ht]
    \centering
    \includegraphics[width=0.8\textwidth]{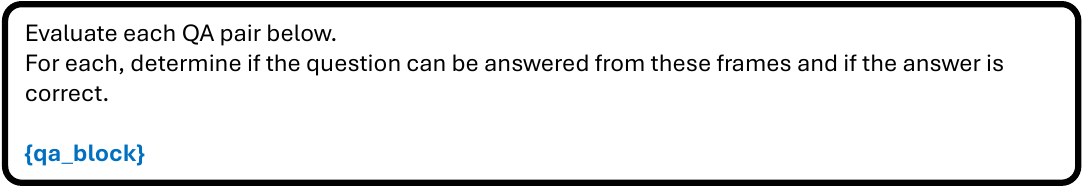}
    \caption{Prompt for GPT Answerability Filtering.}
    \label{fig:qafiltering-gptanswerability}
\end{figure}

\begin{figure}[ht]
    \centering
    \includegraphics[width=0.8\textwidth]{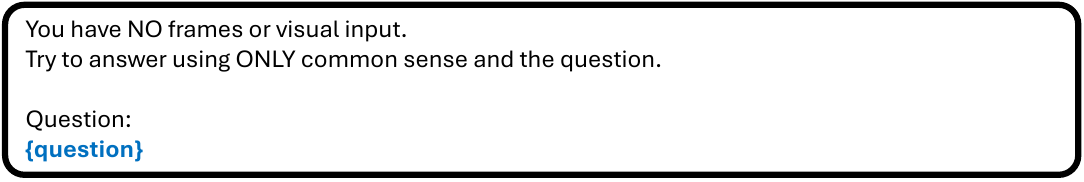}
    \caption{Prompt for GPT Blind Check Filtering.}
    \label{fig:qafiltering-gptblindcheck}
\end{figure}

Figure~\ref{fig:qafiltering-gptanswerability} shows the prompt used to verify whether each generated QA pair is answerable from the provided frames and whether the answer is correct. 
In addition, Figure~\ref{fig:qafiltering-gptblindcheck} presents the prompt used to check whether a question can be answered without visual input, using only common sense and the question itself.

\subsection{LLM Judge Prompt}

\begin{figure}[ht]
    \centering
    \includegraphics[width=0.8\textwidth]{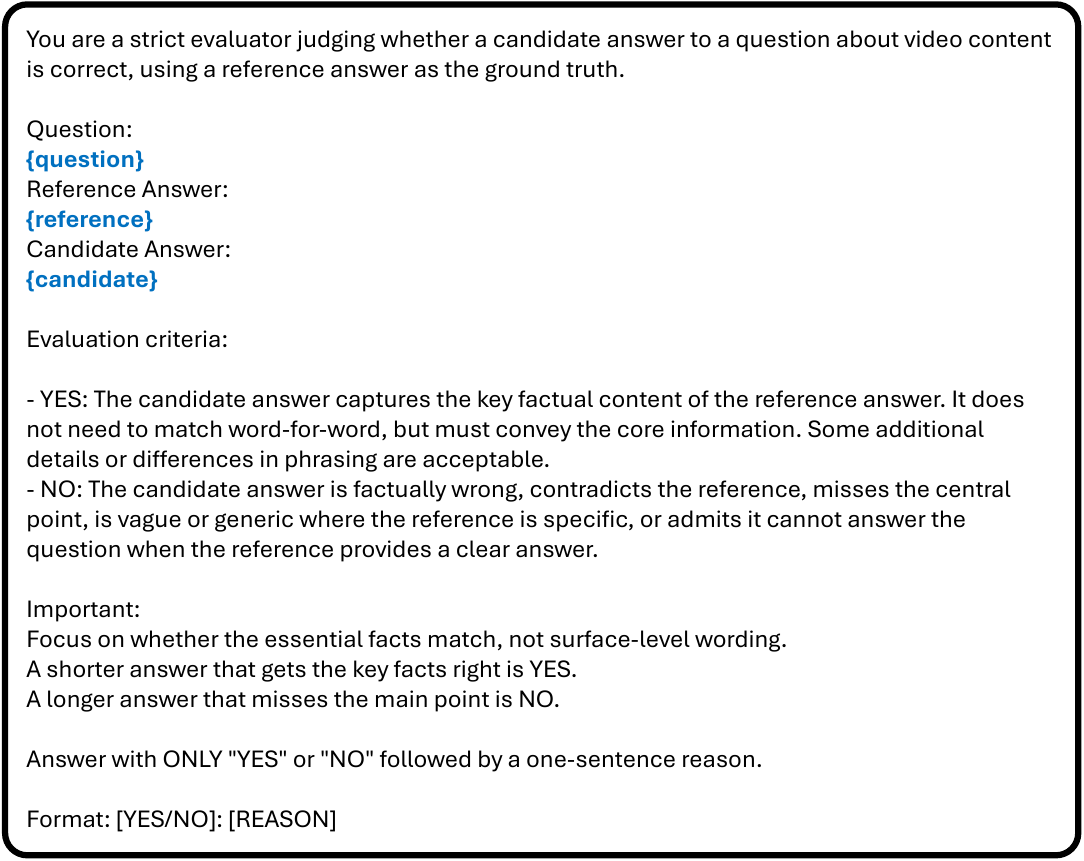}
    \caption{Prompt for LLM Judge.}
    \label{fig:llmjudge}
\end{figure}

Figure~\ref{fig:llmjudge} presents the prompt used by the LLM judge to evaluate whether a candidate answer matches the key factual content of the reference answer.

\subsection{Memory Generation Prompt}

\begin{figure}[ht]
    \centering
    \includegraphics[width=0.8\textwidth]{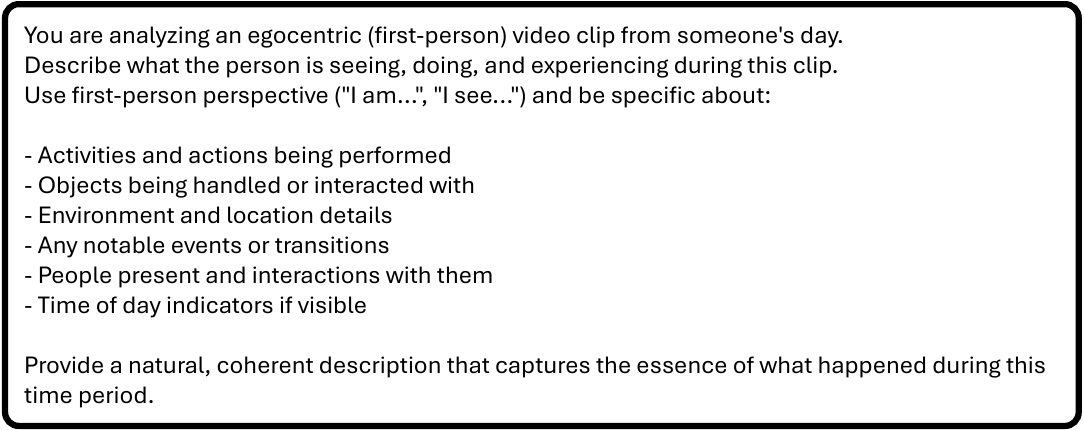}
    \caption{Prompt for Text Clip Memory Generation.}
    \label{fig:memory-textcoarse}
\end{figure}

\begin{figure}[ht]
    \centering
    \includegraphics[width=0.8\textwidth]{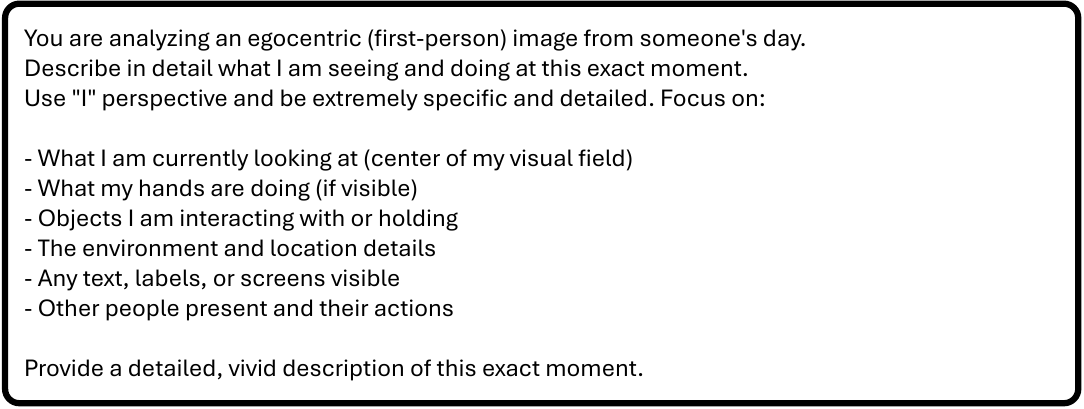}
    \caption{Prompt for Text Key Frame Memory Generation.}
    \label{fig:memory-textfine}
\end{figure}

Figure~\ref{fig:memory-textcoarse} presents the prompt used to generate text-clip memories that summarize what the camera wearer sees, does, and experiences over an egocentric video clip. 
In addition, Figure~\ref{fig:memory-textfine} presents the prompt used to generate text-keyframe memories that describe the camera wearer’s visual experience at a specific moment in detail.

\subsection{QA Answering Prompt}

\begin{figure}[ht]
    \centering
    \includegraphics[width=0.8\textwidth]{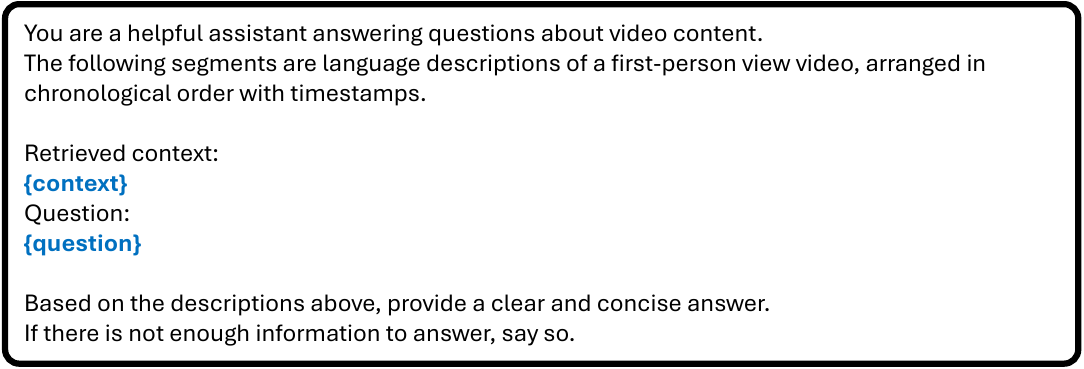}
    \caption{Prompt for Text Key Frame and Text Clip Answering.}
    \label{fig:QAanswering-textfineandcoarse}
\end{figure}

\begin{figure}[ht]
    \centering
    \includegraphics[width=0.8\textwidth]{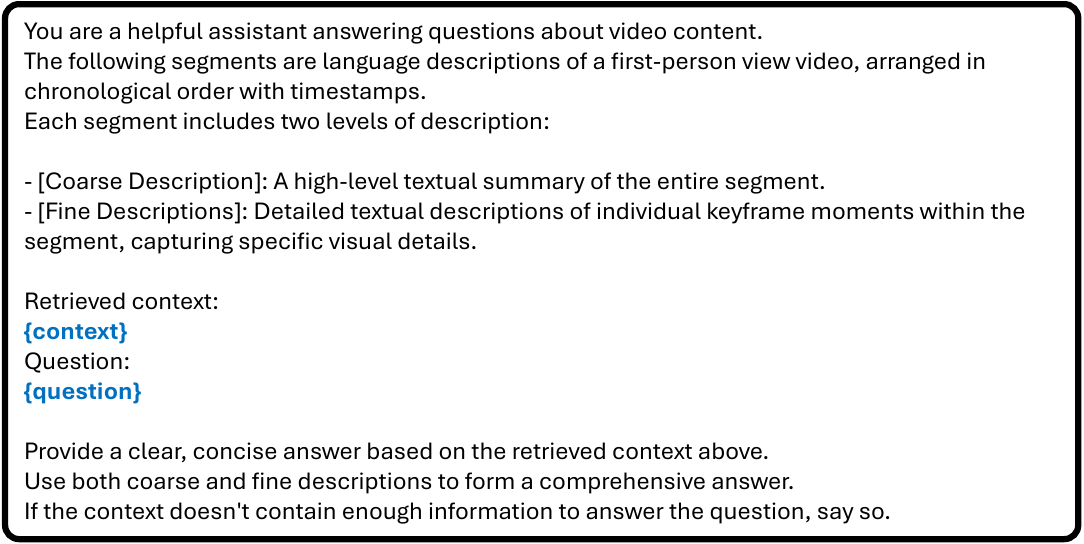}
    \caption{Prompt for Text Combined Answering.}
    \label{fig:QAanswering-textcombined}
\end{figure}

\begin{figure}[ht]
    \centering
    \includegraphics[width=0.8\textwidth]{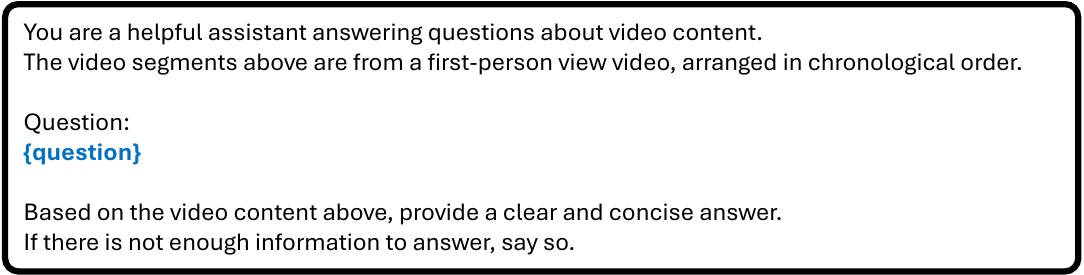}
    \caption{Prompt for Visual Clip Answering.}
    \label{fig:QAanswering-visualcoarse}
\end{figure}

\begin{figure}[ht]
    \centering
    \includegraphics[width=0.8\textwidth]{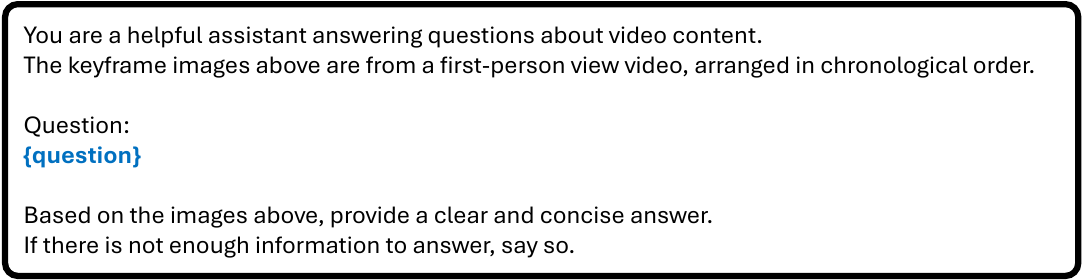}
    \caption{Prompt for Visual Key Frame Answering.}
    \label{fig:QAanswering-visualfine}
\end{figure}

\begin{figure}[ht]
    \centering
    \includegraphics[width=0.8\textwidth]{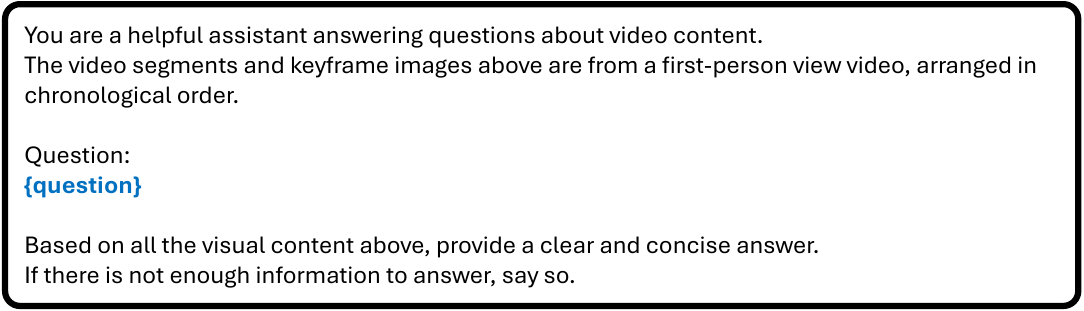}
    \caption{Prompt for Visual Combined Answering.}
    \label{fig:QAanswering-visualcombined}
\end{figure}

Figure~\ref{fig:QAanswering-textfineandcoarse} presents the prompt used to answer questions from retrieved textual descriptions, including either keyframe or clip memory contexts. 
In addition, Figure~\ref{fig:QAanswering-textcombined} presents the prompt used to answer questions from combined textual memories that include both clip-level descriptions and keyframe-level descriptions.
Furthermore, Figure~\ref{fig:QAanswering-visualcoarse} shows the prompt used to answer questions from retrieved video segments, while Figure~\ref{fig:QAanswering-visualfine} shows the prompt used to answer questions from retrieved keyframe images.
Finally, Figure~\ref{fig:QAanswering-visualcombined} presents the prompt used to answer questions from combined visual evidence, including both video segments and keyframe images.

\end{document}